\documentclass{article} 
\PassOptionsToPackage{table}{xcolor} 
\usepackage{iclr2026_conference,times}

\usepackage{amsmath,amsfonts,bm}

\def\eqref#1{equation~\ref{#1}}

\def\1{\bm{1}}

\def\rn{{\textnormal{n}}}

\def\ry{{\textnormal{y}}}

\DeclareMathAlphabet{\mathsfit}{\encodingdefault}{\sfdefault}{m}{sl}
\SetMathAlphabet{\mathsfit}{bold}{\encodingdefault}{\sfdefault}{bx}{n}

\newcommand{\E}{\mathbb{E}}

\newcommand{\KL}{D_{\mathrm{KL}}}

\usepackage{hyperref}
\usepackage{url}
\usepackage{graphicx}
\usepackage{grffile} 

\usepackage{algorithmic}
\usepackage{booktabs}
\usepackage{tikz}
\usepackage[plain]{algorithm}

\usepackage{booktabs}
\usepackage{multirow}
\usepackage{graphicx}
\usepackage{amsthm}
\usepackage[verbose]{wrapfig}   
\usepackage{needspace}
\usepackage[table]{xcolor}
\usepackage{caption}
\usepackage{wrapfig}
\usepackage{xcolor}
\usepackage{tikz}
\usepackage[most]{tcolorbox}   
\usepackage{enumitem}

\newtheorem{proposition}{Proposition}
\newtheorem{corollary}{Corollary}
\newtheorem{definition}{Definition}
\newtheorem{remark}{Remark}
\newtheorem{theorem}{Theorem}
\newtheorem{lemma}{Lemma}
\newtheorem{assumption}{Assumption}
\definecolor{tgThink}{RGB}{140, 26, 17}   
\definecolor{tgPlan} {RGB}{ 93, 14,196}   
\definecolor{tgMem}  {RGB}{191,107, 39}   
\definecolor{tgAct}  {RGB}{ 42,100, 56}   
\definecolor{tgStep} {RGB}{ 55,125,247}   

\newcommand{\Think}[1]{\fld{tgThink}{T}{#1}}
\newcommand{\Plan}[1]{\fld{tgPlan}{P}{#1}}
\newcommand{\Mem}[1]{\fld{tgMem}{M}{#1}}
\newcommand{\Act}[1]{\fld{tgAct}{A}{\textcolor{tgAct}{\ttfamily #1}}}

\usepackage{booktabs}
\usepackage{tabularx}
\usepackage{threeparttable}
\usepackage{xcolor}
\usepackage{array}

\usepackage{pgfplots}
\usepackage{pgfplotstable}

\usetikzlibrary{patterns}
\pgfplotsset{compat=1.18}
\definecolor{before}{HTML}{E8B6CF}
\definecolor{after}{HTML}{AED7D8}

\usepgfplotslibrary{fillbetween}
\pgfplotsset{compat=1.18}

\definecolor{rationale}{RGB}{0,150,190}
\definecolor{uniform}{RGB}{220,65,70}
\definecolor{ppwkd}{HTML}{2E8B57} 
\usetikzlibrary{shapes.geometric}
\definecolor{ppedit}{HTML}{5C8ED6}
\definecolor{ppctl}{HTML}{8A8A8A}
\definecolor{pptko}{HTML}{D9A84E}
\definecolor{ppsty}{HTML}{E0855E}
\definecolor{pptch}{HTML}{5C8ED6}
\definecolor{ppmem}{HTML}{4FB391}
\definecolor{ppact}{HTML}{D9A84E}
\definecolor{ppplan}{HTML}{E0855E}
\definecolor{ppstud}{HTML}{9A9A9A}

\usetikzlibrary{patterns,arrows.meta}
\definecolor{darkgreen}{RGB}{0,100,0}

\title{Teaching Agents in their Own Style: Blackbox Teacher Distillation for Web Agents}

\iclrfinalcopy
\author{Md Farhan Ishmam and Kenneth Marino\\
 Kahlert School of Computing, University of Utah\\
 \texttt{\{farhan.ishmam,kenneth.marino\}@utah.edu} \\[1em]
}

\newcommand{\name}{\textsc{WebKD}}

\begin{document}

\maketitle
\lhead{Preprint} 
\vspace{-24pt} 
\begin{center}
\url{https://web-teach-kd.github.io/}
\end{center}

\begin{abstract}
Frontier web agents using closed models have high inference and computational costs. Smaller open models can be an alternative, but to be performant, they require training supervision from a human or teacher model.
Token-level teacher distillation is not an option for closed teachers as it needs access to the model's logits. The standard approach is to imitate teacher demonstrations offline, but it trains the student on states it has never visited, and in a style it does not write in. We introduce \name, where the teacher is the editor of the student's failures. Given a failed episode, the teacher finds the first point of mistake, and rewrites the student's reasoning traces, harness states, and actions from that point onward. The editing reveals that the teacher's corrections can be more surprising to the student than the student's own text. \name~instead aims to edit minimally. In the data space, the teacher writes each correction in the student's style. In the objective space, the reasoning is weighted by its contribution to the action. A role-asymmetric KL anchor allows the corrected action to move freely while keeping the reasoning near the student's prior. On the TimeWarp benchmark, \name\ students gain about $5.5$-$10.7$pp over behavior cloning, outcome-reward RL, and teacher editing baselines. The performance transfers to unseen versions and to the live web in the OnlineMind2Web benchmark. \name~opens a new way of distilling black-box models to improve the capabilities of open-weight web agents.
\end{abstract}
\vspace{-0.4cm}

\section{Introduction}
\vspace{-0.1cm}
Agents built on the largest proprietary {large language models (LLMs)} now perform remarkably well on web tasks \citep{song2025bearcubs,xue2025onlinemind2web}. Such a task hands the agent a goal and a browser, and
asks it to navigate complex, changing interfaces, plan across tens of steps, and carry
intermediate findings through to a final answer. Deploying these agents for everyday use is
another matter. Frontier models have high latency because they are large reasoning models. The computational cost makes them less viable as agents that run in the background of daily work. Robotics faced a similar problem, which led to compact Vision-Language-Action (VLA) models distilled for real-time control \citep{brohan2023rt2,kim2024openvla}. Web agents also need a small and cheap open-weight model that runs at interactive latency but carries the competence closer to its frontier
teacher.

{This raises the main challenge we address in this paper: \textit{how do we effectively transfer the competence of the black-box, teacher model to the small, more efficient model?} Web tasks can be long-horizon over partially observed pages, and the agent must plan explicitly while reasoning in natural language \citep{xue2025onlinemind2web}. The harness truncates the history to recent observations due to the limited context. So the agent also keeps a scratchpad memory, and what it
writes there, together with its earlier responses, is the only state that persists across steps \citep{dechezelles2025browsergym}
(\S\ref{subsec:background}).}

{Secondly, the outcome reward is a single value per
episode, far too sparse to shape multiple structured decisions with reinforcement learning (RL)
\citep{shao2024deepseekmath,wei2025webagentr1}. Teacher supervision is denser, but the strongest
teachers are black boxes that expose no internals or logits. Every distillation method that consumes the teacher's next-token distribution, \textit{e.g.}, token-level knowledge distillation
(KD), reverse Kullback--Leibler (KL) objectives, and on-policy distillation
\citep{gu2024minillm,agarwal2024onpolicy,lu2025onpolicy}, is therefore unavailable. What a
frontier teacher \emph{can} emit is text. Our research asks \emph{how much of a black-box frontier
teacher's competence can be distilled into a small web agent through text alone}.}

The standard approach is behavior cloning, which involves sampling teacher demonstrations and fine-tuning the student on them offline \citep{kim2016sequence}. For web agents, this is known to help
substantially \citep{he2024openwebvoyager,ishmam2026timewarp}, and it is where our pipeline
begins. But behavior cloning creates two new problems. It trains the student only at states the
\emph{teacher} visits, so at deployment, small errors carry the student into states its
supervision never covered, and the errors compound over the horizon \citep{ross2011dagger}. Secondly, it trains the student on the teacher's surface distribution, unintentionally importing a writing style that the small model cannot comfortably produce. \emph{Is offline imitation really the most a teacher has to offer?}

\begin{figure}[ht]
    \centering
    \includegraphics[width=\linewidth]{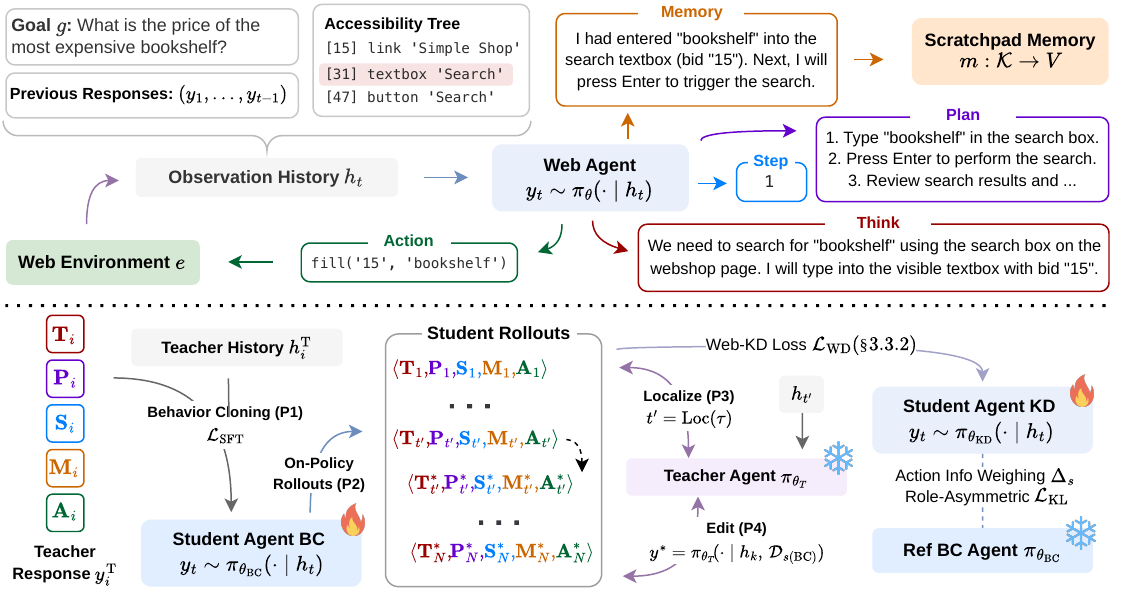}
    \caption{\textbf{Overview of \name.} \textit{(Top)} At step $t$, the harness generates the observation history from the goal, accessibility tree, and previous responses. The web agent emits a five-component response conditioned on the observation history. The \textcolor{tgThink}{\textbf{T}}hink, \textcolor{tgPlan}{\textbf{P}}lan, and \textcolor{tgStep}{\textbf{S}}tep components carry their reasoning and running plan, the \textcolor{tgMem}{\textbf{M}}emory component reads from and writes to the scratchpad memory, and only the \textcolor{tgAct}{\textbf{A}}ction component is executed in the web environment, which returns the next observation. \textit{(Bottom)} The student is behavior-cloned on the teacher's rollouts \textbf{(P1)} and then generates rollouts from the training tasks \textbf{(P2)}. The teacher localizes each failure at $t'$ \textbf{(P3)} and rewrites the faulty trajectory suffix in the student's style \textbf{(P4)}. The resulting corpus $\mathcal{D}$ trains the student under action informativeness weighing, and KL anchored objective of \S\ref{sec:actw}--\S\ref{sec:kl}.}
    \vspace{-1.2em}
    \label{fig:overview}
\end{figure}

The classical answer to this is DAgger \citep{ross2011dagger}, where the expert relabels the states the \emph{learner} visits. However, DAgger corrects only the actions while a web agent's generation is a structured response ({Fig.~\ref{fig:overview}}). With truncated history, the non-action parts of that response form the belief state that guides later decisions. Fixing the click while leaving the reasoning that led to it unchanged, therefore fixes the immediate mistake but not its source. Conversely, on-policy expert correction \citep{lauffer2025oec} gives control to the teacher from the point of failure and lets it finish the episode. So the completed suffix is the teacher's trajectory, which is off-policy in state and style.

An alternative is to use the teacher as an \emph{editor} of the student's own failures. The teacher
reads a failed episode, localizes the earliest erroneous step, and rewrites the student's full
structured response, reasoning, and harness traces included, from that step to the end. Each rewrite is a single-step teacher rollout on a state the student actually visited, so the supervision is on-policy. However, it is not on-policy in style. A frontier teacher's corrections are \textit{surprising} to the student, which are at several times the perplexity of the student's own steps. This gap is especially large in the reasoning components, where perplexity can increase by up to an order of magnitude, while the action itself remains much closer to the student's distribution.


{Under uniform fine-tuning, the loss is largest where the supervision is most foreign, which is the teacher's \emph{wording} rather than its \emph{decisions}.} \name\ attacks this \emph{perplexity premium} by editing the rationale \textit{minimally}. In data space, the teacher writes each correction in the student's own style from in-context exemplars of the student's writing. In objective space, each reasoning span is weighted by how much it raises the student's probability of the corrected action. A role-asymmetric anchor allows the corrected action to move freely while keeping the reasoning close to the student's prior. The student adopts the teacher's decisions in full, while being closer to its own writing style.

Our contributions are as follows:
\begin{itemize}[leftmargin=20pt]
    \item {\textbf{Problem.} We formulate distillation from a \emph{black-box} frontier
    teacher into a small web agent by using teacher-edits and identify a key obstacle, the \emph{on-policy
    perplexity premium} (Defn.~\ref{def:premium}). The premium is the excess surprisal of the student on the teacher supervision, and it concentrates on the reasoning components rather than the action ({Fig~\ref{fig:perplexityPremium}}).}
    
    \item {\textbf{Method.} We introduce \name, a new distillation method that uses the teacher to edit the student's failures without the teacher's logits (Fig.\ref{fig:overview}). The teacher localizes the first erroneous step and rewrites the subsequent reasoning traces, scratchpad memory, and actions in the student's style instead of imitating teacher-style traces (\S\ref{subsec:ourApproachMethodWebKD}). The student then trains under
    a novel objective function that weights each reasoning span by how much it informs the action
    (\S\ref{sec:actw}) and anchors the reasoning to the student's prior
    (\S\ref{sec:kl}).}
    
    \item {\textbf{Evidence.} On the TimeWarp benchmark, \name\ lifts the performance of students by $5.5-10.7$ pp across behavior cloning, outcome-reward RL, and teacher editing baselines using Qwen and Gemma models (Tab.~\ref{tab:main_results}). The gains translate to unseen TimeWarp versions, and the live web benchmark, OnlineMind2Web (Fig. \ref{fig:om2Wresults}). Our ablations vary the source of the edits and styles, training objectives, and hyperparameters (Tables~\ref{tab:edit-style-ablation}, \ref{tab:loss-ablation}, Fig.~\ref{fig:kl-success-rate}).}
\end{itemize}

\vspace{-1em}
\section{Related Work}
\label{sec:relatedWork}
\subsection{Web Agents}

{Web agents, LLM-driven agents that operate a browser on a user's behalf \citep{deng2023mind2web}, are a subcategory of computer-use agents, which perceive rendered user interfaces and act through
UI-level commands \citep{xie2024osworld}. The
benchmark ecosystem spans curated demonstrations \citep{deng2023mind2web}, sandboxed sites
\citep{zhou2024webarena}, live websites \citep{he2024webvoyager}, and unified harnesses
\citep{dechezelles2025browsergym}. TimeWarp \citep{ishmam2026timewarp} additionally varies the
\emph{era} of the interface and tests robustness to temporal UI change. On these suites,
{proprietary frontier LLMs reach strong success rates, but their cost and latency make them impractical as always-on daily drivers.} That gap motivates the small open-weight students this paper targets.}

Training pipelines for such students span imitation and RL. WebRL
\citep{qi2025webrl}, WebAgent-R1 \citep{wei2025webagentr1}, and DigiRL \citep{bai2024digirl}
train with outcome-reward RL and inherit its sparse signal, while OpenWebVoyager
\citep{he2024openwebvoyager} and TimeWarp \citep{ishmam2026timewarp} fine-tune on teacher
demonstrations, the recipe we start from. We differ in what happens after imitation. The
teacher corrects the student's own failures instead of supplying more demonstrations or rewards.
Closest to us is WebCoT \citep{hu2025webcot}, which also distills reconstructed chain-of-thought
rationales. However, the teacher rewrites stored records with template heuristics against observations simulated in the style of WebDreamer \citep{gu2024webdreamer}, and it does not localize the first fatal step before rewriting. \name~edits \textit{actual} failures against the observations the student saw from the localized step in the student's own style, using both the data and training objective.

\vspace{-0.3em}
\subsection{Knowledge Distillation}

{Knowledge distillation transfers a teacher's behavior to a smaller student
\citep{hinton2015distilling}, classically by matching token-level distributions or by
fine-tuning on teacher generations at the sequence level \citep{kim2016sequence}. For LLMs, the
white-box distillation methods sharpen this line of work with reverse-KL objectives \citep{gu2024minillm}, on-policy
distillation that scores \emph{student-sampled} prefixes under the teacher
\citep{agarwal2024onpolicy,lu2025onpolicy}, and skew-divergence stabilizations
\citep{ko2024distillm}. All of them consume the teacher's next-token distribution over a shared
vocabulary. Black-box agents do not expose these, which forces distillation through
text only. This introduces the notion of a perplexity premium (\S\ref{subsec:perplexityPremium}), which has previously been observed with LLM distillation \citep{ren-etal-2024-learn} and is one of our paper's main focuses.}

{A second lineage imitates the expert at the learner's own states. DAgger
\citep{ross2011dagger} showed that relabeling learner-visited states turns quadratic compounding
error into linear error, and LLM-era variants revive the recipe through
turn-level policy interpolation \citep{li2026revisitingdagger}, mid-trajectory expert takeover
\citep{lauffer2025oec}, and iterated fine-tuning on privileged feedback
\citep{choudhury2024leap}. We keep the state-distribution insight but change the label. Our teacher rewrites the full structured response offline instead of relabeling actions online. A third line removes the external teacher. Rejection fine-tuning keeps self-generated successes \citep{zelikman2022star,yuan2023rft}, and on-policy self-distillation conditions the same model on privileged information such as a gold solution \citep{zhao2026opsd}, consensus \citep{li2026uopsd}, or rich feedback \citep{hubotter2026sdpo}, with agentic variants that target failure-relevant spans \citep{yeo2026hintsd,li2026serl}. Self-distillation presupposes the 
competence the model already has, which a 4B student on realistic web tasks typically lacks (Tab. \ref{tab:main_results}). So we keep
an external frontier teacher and hold its supervision close to the student's own states and
style.}

%
%

\vspace{-0.7em}
\section{Methodology}
\label{sec:methodology}
\vspace{-0.3em}
\subsection{Background}
\label{subsec:background}

\paragraph{Web tasks as POMDP.}
A web task is specified by a natural-language goal $g$ on an interactive browser environment $e$ \citep{zhou2024webarena,dechezelles2025browsergym}. Performing a web task is naturally formalized as a finite-horizon partially observable {Markov decision process (POMDP)} \citep{kaelbling1998planning} $\langle \mathcal S, \mathcal A, \mathcal O, \mathbb T, \mathbb T_0, \mathbb O, R \rangle${, with a step budget of $T$ actions per episode}. The
underlying browser state $s_t \in \mathcal S$ contains the full document object model (DOM), session, and site state, and is never observed directly. The initial state is drawn as $s_0 \sim \mathbb T_0$. The agent receives the rendering $o_t = \mathbb O(s_t) \in \mathcal O$, in the form of HTML or {an AXTree}. An LLM policy $\pi_\theta$ generates a structured response $y_t\sim\pi_\theta(\cdot\mid h_t)$ autoregressively, conditioned on the observation history  $h_t$. The browser-level action $a_t \in \mathcal A$, \textit{e.g.}\ \texttt{click('63')}, is deterministically
parsed from $y_t$ and executed, producing $o_{t+1}$ through the effectively deterministic
transition $\mathbb T$. The reward is sparse and terminal: $R \in \{0, 1\}$, assigned once at the end of the episode. The task goal $g$ is rendered into every observation, so the full interaction history is $(o_{1:t},\, y_{1:t-1})$. Carrying several high-dimensional observations (a single AXTree runs to thousands of tokens) overflows the context window, so the history is truncated to the \emph{most recent} observation together with the full response history, $h_t = (o_t,\, y_{1:t-1})$. 

\paragraph{Web harness.}
Between the policy and the browser sits the \emph{harness}, the interface that turns the raw POMDP into an agentic loop. A harness also maintains its internal state across steps, namely the response history $y_{1:t-1}$, which carries the scratchpad memory $s_m$ as the sequence of memory components emitted so far (\S\ref{subsec:scratchpadMemory}). At step $t$, the harness (i) generates the observation history $h_t$, (ii) obtains the structured response $y_t \sim \pi_\theta(\cdot \mid h_t)$, (iii) parses the action $f_\text{parse}(y_t)$, and (iv) executes the action to advance the environment $s_{t+1}=\mathbb T(s_t, a_t)$ and $o_{t+1}=\mathbb O(s_{t+1})$.


\paragraph{Agent response.}
The harness enforces a five-component response,
$y_t = 
\langle\,y^{\texttt{think}}_t,~\allowbreak y^{\texttt{plan}}_t,~\allowbreak y^{\texttt{step}}_t,~\allowbreak y^{\texttt{memory}}_t,~\allowbreak
y^{\texttt{action}}_t\,\rangle = \langle\mathrm{T},\mathrm{P},\mathrm{S},\mathrm{M},\mathrm{A}\rangle $
, generated sequentially (Fig~\ref{fig:overview}) \citep{dechezelles2025browsergym}. \texttt{think} tokens constitute the free-form reasoning grounded in the
current observation, \texttt{plan} tokens form the running, numbered multi-step plan, re-emitted and revised at every step, the \texttt{step} token(s) is a single integer naming the current plan step, \texttt{memory} tokens are the agent's interface to the
scratchpad memory $s_m$ (\S\ref{subsec:scratchpadMemory}), and \texttt{action} tokens form one executable browser command. Only the \texttt{action} tokens ($y^{\texttt{action}}_t=a_t$) interact with the environment. Every other component acts on the harness state alone and enters future contexts as part of the observation history $h_t$. 

We provide additional preliminary discussion on behavior cloning (\S\ref{subsec:BehaviorCloning}), knowledge distillation (\S\ref{subsec:knowledgeDistillation}), and the black-box constraint (\S\ref{subsec:blackBoxConstraint}) in the appendix.

\subsection{Perplexity Premium}
\label{subsec:perplexityPremium}
\textbf{Teacher edits are high-perplexity text for the student.} The perplexity of a response $y$ under a policy $\pi$ with context $x$,
\begin{equation}
\mathrm{PPL}_\pi(y \mid x) \;=\; \exp\!\Big(-\tfrac{1}{|y|}\textstyle\sum_{i}\log \pi\!\left(y_i \mid x,\, y_{<i}\right)\Big),
\label{eq:ppl}
\end{equation}
measures how natural the text is to that policy. $1$ means the policy would have emitted it with certainty, and each unit of log-perplexity is a nat of average per-token surprise. The relative quantity that matters for training is the premium the student pays on its supervision.

\begin{definition}[On-policy perplexity premium]
\label{def:premium}
For a step with student response $y$ and supervision target $y^*$ at the same observation
history $h$, the premium is the excess per-token surprisal, under the behavior-cloned student
$\pi_{\theta_\text{BC}}$ (\S\ref{subsec:BehaviorCloning}), of the supervision relative to the student's own text:
\begin{equation}
\Pi(y^*,\, y)
\;=\;
\frac{1}{|y^*|} \sum_{i} -\log \pi_{\theta_\text{BC}}\!\left(y^*_i \mid h, y^*_{<i}\right)
\;-\;
\frac{1}{|y|} \sum_{i} -\log \pi_{\theta_\text{BC}}\!\left(y_i \mid h, y_{<i}\right).
\end{equation}
\vspace{-10pt}
\end{definition}
\begin{wrapfigure}{r}{0.49\textwidth}
  \centering
  \vspace{-0.8\baselineskip}

\definecolor{ppstud}{HTML}{9A9A9A}
\definecolor{pptch}{HTML}{5C8ED6}

\begin{tikzpicture}[xscale=0.78]
\foreach \gx/\gv in {
  1.609/1.5,
  2.750/2,
  4.359/3,
  5.500/4
} {
  \draw[black!12, line width=0.4pt] (\gx,0) -- (\gx,2.25);
  \node[below, black, font=\tiny] at (\gx,-0.02) {\gv};
}

\node[below, black, font=\tiny] at (0,-0.02) {1};

\draw[black!70, line width=0.6pt] (0,0) -- (0,2.25);
\draw[black!40, line width=0.4pt] (0,0) -- (5.5,0);

\foreach \ry/\rn/\sx/\wx/\tx/\tv in {
  1.95/think/1.298/1.558/4.240/2.91,
  1.40/plan/0.200/0.816/2.395/1.83,
  0.85/memory/0.368/0.811/2.020/1.66,
  0.30/action/0.127/0.103/0.194/1.05
} {
  \draw[black!25, line width=1.4pt] (\sx,\ry) -- (\tx,\ry);
  \fill[ppstud] (\sx,\ry) circle (2.1pt);
  \fill[ppwkd]  (\wx,\ry) circle (2.1pt);
  \fill[pptch]  (\tx,\ry) circle (2.1pt);

  \node[right, black, font=\tiny]
    at (\tx+0.07,\ry) {\tv};

  \node[left, black, font=\scriptsize]
    at (-0.12,\ry) {\texttt{\rn}};
}

\fill[ppstud] (-0.10,2.70) circle (2.1pt);
\node[right, black, font=\tiny]
  at (-0.03,2.70) {student's text ($\approx 1.13$)};

\fill[pptch] (3.0,2.70) circle (2.1pt);
\node[right, black, font=\tiny]
  at (3.09,2.70) {teacher's edits ($\approx 1.90$)};

\fill[ppwkd] (-0.10,2.44) circle (2.1pt);
\node[right, black, font=\tiny]
  at (-0.03,2.44) {\vspace{1em}\name's corpus ($\approx 1.25$)};

\node[below, black, font=\tiny]
  at (2.75,-0.28) {per-token perplexity (log)};
\end{tikzpicture}

  \caption{The perplexity premium by response components against the student's continuations, the teacher's raw corrections, and the \name's corpus with in-context edits.} 
  
  \label{fig:perplexityPremium}
  \vspace{-0.9\baselineskip}
\end{wrapfigure}
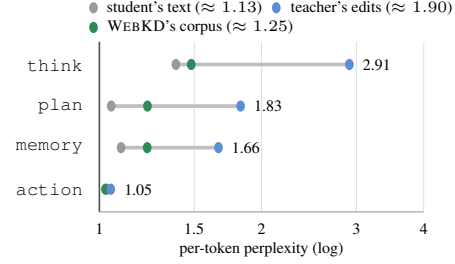

This premium is one of the obstacles to distilling a frontier teacher through text. As shown in Fig. \ref{fig:perplexityPremium}, the teacher edits are high-perplexity for the student, while \name's corpus~stays close to the student's own continuations. 
{Per-token cross-entropy is the surprisal inside Eq.~\ref{eq:ppl}, and under uniform SFT, the loss is largest on the tokens most foreign to the student. On raw teacher edits, those are mostly the rationale tokens. The action tokens that fix the failure carry only a few
percent of the loss. The surprising spans mix the teacher's style
with its content and reasoning. What it shows is
where uniform SFT spends its loss, which motivates the objective of Eq.\ref{eq:kl}. Any method that trains on teacher edits must address the premium.} Our method attacks it from both sides. \emph{In data space}, before training, the correction is written in the student's own style (P4 of \S\ref{sec:methodology}). \emph{In objective space}, during training, the learning signal is concentrated on the tokens that inform the action (\S\ref{sec:actw}), and the rationale is anchored to the student's own prior{\color{black}, which damps the premium the student absorbs and caps it per token once $\beta$ exceeds the threshold of Corollary~\ref{cor:beta}} (\S\ref{sec:kl}).



\subsection{Web Knowledge Distillation (\name)}
\label{subsec:ourApproachMethodWebKD}

The core of \name\ is to correct the student's decisions at the states it visited, rewriting only the steps from the failure onward. \emph{How can the student policy be improved while minimizing the student's surprisal to teacher edits?} \name~starts from a student agent fine-tuned to replicate the teacher (\textbf{P1}). We then collect rollouts from the fine-tuned student agent and keep its failures (\textbf{P2}).
The edit itself is a two-stage teacher intervention: \textit{where did the student fail?} (\textbf{P3}), and \textit{how should the teacher correct it?} (\textbf{P4}). The corpus of corrected trajectories trains the student under a premium-aware objective: action-informative token weighting (\S\ref{sec:actw}) and a role-asymmetric anchor to the student's prior (\S\ref{sec:kl}). Figure~\ref{fig:overview} gives an overview, and Algorithm~\ref{alg:webkd} states the full method.  We characterize in closed form {the per-token target of} the \name~objective
    (Proposition~\ref{prop:target}) and bound the student's failure rate by the teacher's
    localization and correction errors {plus the student's own disagreement with the corrected process} (Theorem~\ref{thm:bound}).


\definecolor{hdStudent}{RGB}{ 90, 90, 90}
\definecolor{hdRaw}    {RGB}{150, 70, 60}
\definecolor{hdOurs}   {RGB}{  0,105,120}

\newcommand{\flabel}[2]{\textcolor{#1}{$\langle\mathrm{#2}\rangle$}}

\newlength{\fldw}
\newcommand{\fld}[3]{\par\noindent\hangindent=\fldw\hangafter=1%
  \makebox[\fldw][l]{\flabel{#1}{#2}}#3}

\newcommand{\pitem}{\newline}
\newcommand{\psub}{\newline\hspace*{0.8em}--\,}

\definecolor{wtText}{RGB}{110,110,110}
\newcommand{\wt}[1]{{\color{wtText}\sffamily\tiny[#1]}}

\tcbset{rollpanel/.style={enhanced, colback=white, colframe=#1,
  colbacktitle=#1, coltitle=white, boxrule=0.6pt, arc=1.5pt,
  left=3pt, right=3pt, top=2pt, bottom=2pt, toptitle=1pt, bottomtitle=1pt,
  fonttitle=\sffamily\bfseries\fontsize{8}{10}\selectfont, halign title=center,
  before title=\strut,
  fontupper=\rmfamily\scriptsize, halign=flush left, 
  before upper={\setlength{\parskip}{2.5pt}%
    \settowidth{\fldw}{$\langle\mathrm{M}\rangle$\enspace}}}}

\begin{figure}[t]
\centering
\begin{tcbraster}[raster columns=3, raster equal height,
                  raster column skip=4pt, raster force size=false]
%
\begin{tcolorbox}[rollpanel=hdStudent, add to width=-17pt, title={Student's Step}]
\Think{Bought Welch's; code B644DDC66C. Next is the Mojito mix, so I
click on the search box (bid 15).}
\Plan{1.~Search Mojito mix
  \psub click box, type, Enter
  \pitem 2.~Open first result, buy
  \psub Buy Now, note code
  \pitem 3.~Search Italian Basil Pesto
  \pitem [\ldots]
  \pitem 5.~Send all codes to user}
\Mem{Clicking the search box to order the Mojito mix.}
\Act{click('15')}
\end{tcolorbox}
%
\begin{tcolorbox}[rollpanel=hdRaw, add to width=-9pt, title={Raw Teacher Edit}]
\Think{Order for Welch's Orange Pineapple Juice is confirmed
(B644DDC66C). I will go back to the shop home page.}
\Plan{1.~Record Welch's code
  \pitem 2.~Go to shop home
  \pitem 3.~Search Stirrings Mojito mix
  \pitem 4.~Buy it, record code
  \pitem 5--7.~Repeat 2--4 for Italian Basil Pesto
  \pitem 8.~Send all codes to user}
\Mem{Welch's confirmation code: B644DDC66C.}
\Act{goto("/shop")}
\end{tcolorbox}
%
\begin{tcolorbox}[rollpanel=hdOurs, add to width=26pt, title={Style-Conditioned Edit (\name)}]
\Think{\wt{.10} Welch's confirmation shows code B644DDC66C.
\wt{.90} I will return to the shop search page for the next product.}
\Plan{\wt{.80} 1.~Record Welch's code
  \pitem \wt{.40} 2.~Search Mojito mix
  \psub go to shop, search, open result
  \psub \wt{1.0} Buy Now, record code
  \pitem \wt{.70} 3.~Search Italian Basil Pesto
  \psub (same steps)
  \psub \wt{.60} Buy Now, record code
  \pitem \wt{.50} 4.~Send all codes to user}
\Mem{\wt{.30} Welch's code is B644DDC66C.
\wt{.20} Next: search Stirrings Mojito mix.}
\Act{goto("/shop")}
\end{tcolorbox}
\end{tcbraster}
\caption{\textbf{Edits for the same student step.} The raw teacher edit replaces
the student's action but rewrites the plan in its own style. The style-conditioned edit keeps the student's nested plan format. Values in
brackets give the span weight $v_s$.
\flabel{tgThink}{\textbf{T}}: think, \flabel{tgPlan}{\textbf{P}}: plan,
\flabel{tgMem}{\textbf{M}}: memory, \flabel{tgAct}{\textbf{A}}: action.}
\label{fig:rollout-edit}
\vspace{-2em}
\end{figure}
\vspace{-5pt}
\paragraph{P1: Behavior cloning on teacher rollouts.}
\name\ first replicates the teacher rollouts on the training tasks via behavior cloning at the sequence level. For the teacher $\pi_{\theta_\text{T}}$ acting as the agent under the same harness, the base model $\pi_{\theta_0}$ is fine-tuned on the teacher demonstration trajectories $\mathcal{D}_\text{T}=\{\tau_n\}_{n=1}^N$, where each $\tau_n = (h_1, y_1, \ldots, h_T, y_T)$ is a closed-loop teacher rollout, $y_i \sim \pi_{\theta_\text{T}}(\cdot \mid h_i)$, kept only if its terminal reward is $R(\tau_n)=1$. Training uses the uniform BC objective $\mathcal{L}_\text{BC}$ of
Eq.~\ref{eq:sft} to produce $\pi_{\theta_\text{BC}}$, which is then frozen and reused in the following phases. 
\vspace{-5pt}
\paragraph{P2: Student on-policy rollout.}
We collect on-policy rollouts from the behavior-cloned student on the training tasks. Each rollout is generated by the agentic loop of \S\ref{subsec:background}, every response drawn from the student's own policy at a context the student itself reached, $y_t \;\sim\; \pi_{\theta_\text{BC}}(\cdot \mid h_t)$ and we write $\tau \sim \pi_{\theta_\text{BC}}$ for the induced trajectory distribution. $\mathcal{D}_\text{BC}$ collects these rollouts. The failed trajectories $\mathcal{D}_{f(\text{BC})} = \{\tau \in \mathcal{D}_\text{BC} : R(\tau) < 1\}$ become the substrate for editing, and the solved episodes are kept as a \emph{style pool} $\mathcal{D}_{s(\text{BC})} = \{\tau \in \mathcal{D}_\text{BC} : R(\tau) = 1\}$, mined for in-context exemplars in P4. Every observation in $\mathcal{D}_{f(\text{BC})}$ has been visited by the student, which places the teacher's supervision of the following phases within the student's own state distribution.

\paragraph{P3: Failure localization.}
For each failed trajectory, the teacher first localizes the error to prevent editing the correct steps. We establish that web trajectories carry enough structure for the ``point of failure'' to be a well-defined property of the trajectory (\S\ref{subsec:p3failureLocalizationDetails}). The teacher call is an \emph{estimator} of $t^\star$, reads the full trajectory $\tau\in\mathcal{D}_{f(\text{BC})}$ and returns an index, $t' = \mathrm{Loc}(\tau) \in \{1, \ldots, T\}$, the \emph{earliest} step whose action sets the trajectory on the path to failure. Localization splits the trajectory into {a replayed prefix} $\tau_p=(h_1,y_1,\dots,h_{t'-1},y_{t'-1})$, which {is kept verbatim}, and a faulty suffix $\tau_s=(h_{t'},y_{t'},\dots,h_{T},y_{T})$, which will be rewritten. 

\paragraph{P4: Teacher editing in the student's style.}
{After localization, the teacher corrects the structured response of every step in the faulty
suffix. Each correction is conditioned on the observation the student saw at that step,
re-rendered by replaying the student's original actions in the deterministic container, so the teacher sees exactly what the student saw. The correction query also carries the student's response style \textit{in context}.}
The corpus targets are
\begin{equation}
y_k^* \;=\; y_k \quad \forall k < t'~(\text{replay}),
\qquad\quad
y_k^* \;\sim\; \pi_{\theta_T}\!\left(\cdot \mid h_k,\, \mathcal D_{s(\text{BC})}\right) \quad \forall k \ge t'~(\text{edit}),
\label{eq:edit}
\end{equation}
where each edit is a $1$-step rollout of the teacher policy at the \emph{student's} context.
$\pi_T$ is conditioned on the observation history $h_k$ the student saw (the same rendered goal,
AXTree, and response history) and on in-context exemplars from the student's style pool
$\mathcal{D}_{s(\text{BC})}$, one independent call per step. The edited structured response
$y^*_k$ comprises the edited tokens $\langle\mathrm{T}^*,\mathrm{P}^*,\mathrm{S}^*,\mathrm{M}^*,\mathrm{A}^*\rangle$ (Details in \S\ref{subsec:teacherEditDetails}).

\subsubsection{Action-informative token weighting}
\label{sec:actw}

{P4 first lowers the premium in data space by having the teacher write in the student's style. It cannot lower it to zero without giving up the correction. Some premium must be paid, and Fig.~\ref{fig:perplexityPremium} shows that on raw teacher edits, it falls almost entirely on the rationale components. The question for the objective is therefore \emph{which} rationale tokens are worth their premium. 

We ask, \emph{which tokens carry the learning signal?} A rationale span is worth its premium to the extent that it changes the action, so Eq.\ref{eq:actw} ranks the spans of each step by their measured effect on the student's probability of the corrected action. Second, \emph{how far the policy may move?} Eq.\ref{eq:kl} adds a role-asymmetric anchor that damps how far the rationale moves toward the teacher's phrasing while the action follows the teacher in full. Together, the objective pays the premium mainly where it buys the action.} We weigh each rationale span by how much it contributes to the decision that follows.

Let $s$ be a non-action segment, a sentence or clause of the edited rationale
$\langle\mathrm{T}^*,\mathrm{P}^*,\mathrm{M}^*\rangle$. Its leave-one-out contribution to the teacher's corrected action, measured under the \emph{student} is
\begin{equation}
{\Delta_s \;=\; \log P_{\pi_{\theta_\text{BC}}}\!\left(a^* \mid h, y^*_{<a}\right)
\;-\; \log P_{\pi_{\theta_\text{BC}}}\!\left(a^*\mid h, y^*_{<a} \setminus s\right)},\quad s \subset \langle\mathrm{T}^*,\mathrm{P}^*,\mathrm{M}^*\rangle,
\label{eq:actw}
\end{equation}
where $a^*=\langle\mathrm{A}^*\rangle$ denotes the teacher-edited action tokens, $h$ is the observation history,
{$y^*_{<a} = \langle\mathrm{T}^*,\mathrm{P}^*,\mathrm{S}^*,\mathrm{M}^*\rangle$ is the part of the edited response that precedes the action, so both terms are teacher-forced log-probabilities of the action tokens, and $y^*_{<a} \setminus s$ is that prefix with segment $s$ deleted.} As $\Delta_s$ is the
\emph{student's own} credit assignment over the teacher's rationale, it needs no teacher logits and is computable in the black-box setting. Perplexity says how foreign a span is to the student. $\Delta_s$ says how much the span raises the student's probability of the corrected action, and two spans can be equally foreign while differing in this. The weighting in \S\ref{subsec:WebKDLossDetails} ranks the spans of a step by $\Delta_s$, so{\color{black}, outside the dead zone of Eq.~\ref{eq:actw-map} and when these spans are unique,} the span that moves the action the most keeps full weight, and the span that moves it the least drops to the floor. Hence, the premium is paid mainly where it changes the decision rather than where it only changes the wording or style. Details on the action informativeness mapped to the \name~loss have been provided in \S\ref{subsec:WebKDLossDetails}.

\begin{table*}[t]
\centering
\caption{Success rate (\%) on the TimeWarp test split, in domain on v6 and on the five held-out versions v1--v5. Results are averaged over three runs with the standard error as the subscript.}
\label{tab:main_results}
\resizebox{\textwidth}{!}{%
\begin{tabular}{llrrrrrr}
\toprule
\textbf{Model} & \textbf{Baseline} &
\multicolumn{1}{c}{\textbf{v1}} & \multicolumn{1}{c}{\textbf{v2}} & \multicolumn{1}{c}{\textbf{v3}} &
\multicolumn{1}{c}{\textbf{v4}} & \multicolumn{1}{c}{\textbf{v5}} & \multicolumn{1}{c}{\textbf{v6 (seen)}} \\
\midrule

{\cellcolor{gray!15}--} & {\cellcolor{gray!15}Teacher (GPT-5.5)}
& {\cellcolor{gray!15}$71.8_{\pm 4.43}$}
& {\cellcolor{gray!15}$71.8_{\pm 4.43}$}
& {\cellcolor{gray!15}$68.0_{\pm 4.60}$}
& {\cellcolor{gray!15}$68.0_{\pm 4.60}$}
& {\cellcolor{gray!15}$69.9_{\pm 4.52}$}
& {\cellcolor{gray!15}$70.9_{\pm 4.58}$}
\\
\midrule

\multirow{10}{*}{\shortstack[l]{Qwen 3.5\\4B}}
& Zero-Shot
& $1.0_{\pm 0.57}$
& $1.0_{\pm 0.57}$
& $0.3_{\pm 0.31}$
& $1.6_{\pm 0.71}$
& $0.0_{\pm 0.00}$
& $1.0_{\pm 0.57}$
\\\cmidrule{2-8}

& Behavior Cloning
& $10.7_{\pm 1.76}$
& $9.4_{\pm 1.66}$
& $9.7_{\pm 1.68}$
& $10.7_{\pm 1.76}$
& $5.2_{\pm 1.26}$
& $9.7_{\pm 1.68}$
\\

& Behavior Cloning (TW)
& $27.5_{\pm 2.54}$
& $28.5_{\pm 2.57}$
& $27.8_{\pm 2.55}$
& $26.2_{\pm 2.50}$
& {$28.2_{\pm 2.56}$}
& $29.8_{\pm 2.60}$
\\\cmidrule{2-8}

& m-GRPO
& {$30.1_{\pm 2.62}$}
& {$27.2_{\pm 2.54}$}
& {$26.2_{\pm 2.50}$}
& {$28.2_{\pm 2.56}$}
& {$30.1_{\pm 2.61}$}
& {$30.4_{\pm 2.62}$}
\\\cmidrule{2-8}

& Self/Student Edit
& $25.2_{\pm 2.47}$
& $19.4_{\pm 2.25}$
& $27.2_{\pm 2.53}$
& $24.3_{\pm 2.56}$
& $23.3_{\pm 2.61}$
& $23.6_{\pm 2.62}$
\\

& Off-policy Takeover
& $24.4_{\pm 2.45}$
& $24.3_{\pm 2.44}$
& $25.2_{\pm 2.48}$
& $23.3_{\pm 2.41}$
& $27.2_{\pm 2.54}$
& $25.6_{\pm 2.49}$
\\

& Student Re-style Edit
& $28.8_{\pm 2.58}$
& $23.6_{\pm 2.42}$
& $23.7_{\pm 2.43}$
& $26.5_{\pm 2.52}$
& $28.3_{\pm 2.57}$
& $26.9_{\pm 2.53}$
\\\cmidrule{2-8}

& \cellcolor{cyan!06}\name
& \cellcolor{cyan!06}{$\mathbf{37.9}_{\pm 2.77}$}
& \cellcolor{cyan!06}{$\mathbf{36.3}_{\pm 2.73}$}
& \cellcolor{cyan!06}{$\mathbf{35.3}_{\pm 2.72}$}
& \cellcolor{cyan!06}{$\mathbf{38.8}_{\pm 2.77}$}
& \cellcolor{cyan!06}{$\mathbf{36.9}_{\pm 2.74}$}
& \cellcolor{cyan!06}{$\mathbf{41.1}_{\pm 2.80}$}
\\

\midrule

\multirow{10}{*}{\shortstack[l]{Qwen 3.5\\9B}}
& Zero-Shot
& $0.0_{\pm 0.00}$
& $0.3_{\pm 0.31}$
& $1.6_{\pm 0.71}$
& $1.3_{\pm 0.65}$
& $0.0_{\pm 0.00}$
& $0.6_{\pm 0.44}$
\\\cmidrule{2-8}

& Behavior Cloning
& $12.3_{\pm 1.87}$
& $10.0_{\pm 1.71}$
& $6.8_{\pm 1.43}$
& $11.7_{\pm 1.82}$
& $4.9_{\pm 1.23}$
& $12.3_{\pm 1.87}$
\\

& Behavior Cloning (TW)
& $25.9_{\pm 2.49}$
& $24.3_{\pm 2.44}$
& $22.0_{\pm 2.36}$
& $25.2_{\pm 2.47}$
& $26.2_{\pm 2.50}$
& $25.2_{\pm 2.47}$
\\\cmidrule{2-8}

& m-GRPO
& $26.9_{\pm 2.52}$
& $27.8_{\pm 2.54}$
& $28.8_{\pm 2.58}$
& $31.1_{\pm 2.62}$
& $31.4_{\pm 2.65}$
& $26.5_{\pm 2.51}$
\\\cmidrule{2-8}

& Self/Student Edit
& $22.3_{\pm 2.37}$
& $23.2_{\pm 2.40}$
& $22.7_{\pm 2.38}$
& $25.7_{\pm 2.49}$
& $22.7_{\pm 2.38}$
& $23.0_{\pm 2.39}$
\\

& Off-policy Takeover
& $28.4_{\pm 2.57}$
& $25.4_{\pm 2.48}$
& $29.4_{\pm 2.59}$
& $30.1_{\pm 2.61}$
& $27.3_{\pm 2.53}$
& $25.0_{\pm 2.46}$
\\

& Student Re-style Edit
& $17.3_{\pm 2.15}$
& $19.5_{\pm 2.25}$
& $18.7_{\pm 2.22}$
& $14.4_{\pm 2.00}$
& $16.4_{\pm 2.10}$
& $20.2_{\pm 2.28}$
\\\cmidrule{2-8}

& \cellcolor{cyan!06}\name
& \cellcolor{cyan!06}{$\mathbf{34.6}_{\pm 2.71}$}
& \cellcolor{cyan!06}{$\mathbf{32.3}_{\pm 2.66}$}
& \cellcolor{cyan!06}{$\mathbf{32.4}_{\pm 2.66}$}
& \cellcolor{cyan!06}{$\mathbf{34.5}_{\pm 2.70}$}
& \cellcolor{cyan!06}{$\mathbf{34.9}_{\pm 2.71}$}
& \cellcolor{cyan!06}{$\mathbf{34.6}_{\pm 2.71}$}
\\\midrule

\multirow{10}{*}{\shortstack[l]{Gemma 4\\E4B}}
& Zero-Shot
& $1.9_{\pm 0.78}$
& $7.8_{\pm 1.53}$
& $7.8_{\pm 1.53}$
& $1.9_{\pm 0.77}$
& $3.9_{\pm 1.10}$
& $2.6_{\pm 0.90}$
\\\cmidrule{2-8}

& Behavior Cloning
& $17.5_{\pm 2.16}$
& $16.5_{\pm 2.11}$
& $18.4_{\pm 2.20}$
& $15.5_{\pm 2.06}$
& $16.5_{\pm 2.11}$
& $19.4_{\pm 2.25}$
\\

& Behavior Cloning (TW)
& $28.2_{\pm 2.56}$
& $28.2_{\pm 2.56}$
& $32.0_{\pm 2.65}$
& $30.1_{\pm 2.61}$
& $28.8_{\pm 2.58}$
& $29.4_{\pm 2.58}$
\\\cmidrule{2-8}

& m-GRPO
& $16.5_{\pm 2.11}$
& $14.6_{\pm 2.01}$
& $17.5_{\pm 2.16}$
& $18.4_{\pm 2.20}$
& $14.6_{\pm 2.01}$
& $20.7_{\pm 2.30}$
\\\cmidrule{2-8}

& Self/Student Edit
& $22.3_{\pm 2.37}$
& $24.3_{\pm 2.44}$
& $21.4_{\pm 2.33}$
& $22.3_{\pm 2.37}$
& $21.7_{\pm 2.34}$
& $23.0_{\pm 2.39}$
\\

& Off-policy Takeover
& $29.8_{\pm 2.60}$
& $28.2_{\pm 2.56}$
& $29.1_{\pm 2.58}$
& $30.1_{\pm 2.61}$
& $33.0_{\pm 2.66}$
& $34.3_{\pm 2.70}$
\\

& Student Re-style Edit
& $20.4_{\pm 2.29}$
& $25.2_{\pm 2.47}$
& $23.3_{\pm 2.40}$
& $24.3_{\pm 2.45}$
& $23.3_{\pm 2.40}$
& $28.2_{\pm 2.56}$
\\\cmidrule{2-8}

& \cellcolor{cyan!06}\name
& \cellcolor{cyan!06}{$\mathbf{37.2}_{\pm 2.75}$}
& \cellcolor{cyan!06}{$\mathbf{34.0}_{\pm 2.69}$}
& \cellcolor{cyan!06}{$\mathbf{37.9}_{\pm 2.76}$}
& \cellcolor{cyan!06}{$\mathbf{36.6}_{\pm 2.74}$}
& \cellcolor{cyan!06}{$\mathbf{38.8}_{\pm 2.77}$}
& \cellcolor{cyan!06}{$\mathbf{39.8}_{\pm 2.78}$}
\\

\bottomrule
\end{tabular}%
}
\vspace{-10pt}
\end{table*}
\vspace{-5pt}
\subsubsection{Role-asymmetric Anchoring and the \name~Objective}
\label{sec:kl}

The weighting decides which spans contribute to the action, but does not bound the magnitude of their contribution. A decisive span in $\langle\mathrm{T}^*\rangle$ keeps full weight, and fitting it under the weighed SFT objective asks the student to fully absorb that span's influence. As this could cause SFT to move the policy further than intended, we add a KL penalty \citep{hinton2015distilling} toward $\pi_{\text{ref}} \equiv \pi_{\theta_\text{BC}}$, the frozen checkpoint of each run. The KL acts on the rationale tokens $\langle \mathrm{T},\mathrm{P},\mathrm{M}\rangle$  only, \textit{i.e.}, {\color{black}it pulls these tokens toward the student's prior and leaves the non-rationale $\langle \mathrm{A}, \mathrm{S}\rangle$ tokens free to drift.} With $\beta \ge 0$ the anchor strength and $\gamma_i \in [0,1]$ a per-token role mask, the \name\ objective is
\begin{align}
\mathcal{L}_\text{WD} \;=\; \underbrace{\frac{\sum_{i \in T} w_i\, \mathrm{CE}_i}{\sum_{i \in T} w_i}}_{\mathcal{L}_{\text{SW}}} \;+\; \beta \cdot
\underbrace{\frac{\sum_{i \in T} \gamma_i \, \mathrm{kl}_i}{\sum_{i \in T} \gamma_i}}_{\mathcal{L}_\text{KL}}
\label{eq:kl}
\end{align}
where $d_i:= \log \pi_{\text{ref}} -\log \pi_\theta(y^*_i)$, $\mathrm{kl}_i := e^{d_i} - {d_i} - 1$, and $\pi_{\theta_{\mathrm{WD}}}$ is the student produced. 
We note that $\gamma$ is \emph{role-asymmetric}. $\gamma_i = 1$ on the $\langle\mathrm{T}, \mathrm{P},\mathrm{M}\rangle$ tokens and $\gamma_i \approx 0$ on
$\langle\mathrm{A},\mathrm{S}\rangle$, so the student is free to adopt the
teacher's corrected decisions while being held near its prior on deliberation. Normalizing by
$\sum_i \gamma_i$ makes $\beta$ a penalty \emph{per anchored token}, comparable across masks
that cover different numbers of tokens. Because $\mathrm{kl}_i$ is evaluated on corpus tokens
rather than on samples from $\pi_\theta$, it is a pointwise anchor penalty on the target
sequence. Details on the intuition of update in \S\ref{subsec:theoreticalIntuition}.

\section{Experiments}
\label{sec:experiments}
We organize our experiments around four research questions:
\textbf{RQ1 (performance).} Does \name's teacher edit distillation outperform
behavior cloning, outcome-reward RL, or other methods of teacher edit?
\textbf{RQ2 (transfer).} Do the gains survive interface versions never seen in training, and the
live web?
\textbf{RQ3 (edit-style).} Who should edit the student's decisions and in whose style?
\textbf{RQ4 (objective).} Which tokens should the loss emphasize, and how should the policy be anchored?

\subsection{Experimental setup}
\label{sec:setup}
We describe our setup concisely here and include additional details in the Appendix (\S\ref{sec:experimentalDetails}).

\textbf{Models, Harness, Baselines.} We compare latest open-weight models: Qwen-3.5 4B and 9B \citep{qwen35blog}, and Gemma 4 E4B \citep{team2026gemma}, in BrowserGym with the AgentLab harness \citep{dechezelles2025browsergym}. We evaluate (a) zero-shot, (b) vanilla behavior cloning on actions, and the TimeWarp approach on the full response \citep{ishmam2026timewarp}, (c) outcome-based RL baseline using multi-turn Group Relative Policy Optimization (m-GRPO) \citep{wei2025webagentr1}, and (d) black-box distillation baselines on student self-editing, off-policy takeover with the teacher completing the trajectory, the teacher prompting the student to restyle the edits, and (e) our method, \name. The average results of three seeds are reported with the standard error.

\textbf{Benchmarks.} We evaluate on: (a) TimeWarp (TW) \citep{ishmam2026timewarp}, a containerized web benchmark with websites from different eras of the internet, and (b) OnlineMind2Web (OM2W) \citep{xue2025onlinemind2web}, a live web benchmark with realistic tasks using real user workflows. 




\begin{figure*}[t]
\centering
\begin{minipage}[t]{0.48\linewidth}
    \centering
    \includegraphics[width=\linewidth]{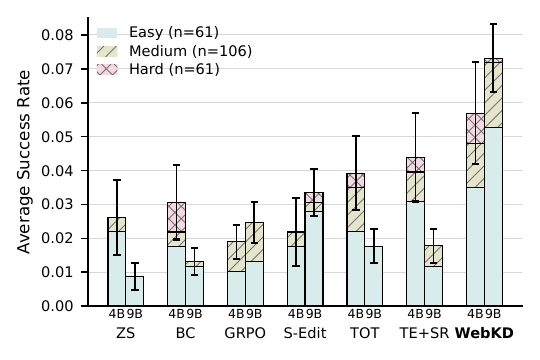}
    \caption{Success rate on Online-Mind2Web for the Qwen-3.5 4B and 9B students. The contributions are stacked across easy, medium, and hard categories.}
    \label{fig:om2Wresults}
\end{minipage}
\hfill
\begin{minipage}[t]{0.48\linewidth}
    \centering
\pgfplotstableread[row sep=\\]{
    idx  y        err     \\
    0    0.3732   0.0270  \\
    1    0.4110   0.0275  \\
    2    0.3743   0.0276  \\
    3    0.2190   0.0276  \\
}\rationaledata

\pgfplotstableread[row sep=\\]{
    idx  y        err     \\
    0    0.3732   0.0270  \\
    1    0.3657   0.0274  \\
    2    0.3463   0.0271  \\
    3    0.2654   0.0252  \\
}\uniformdata

\begin{tikzpicture}
\begin{axis}[
    width=\linewidth,
    height=0.67\linewidth,
    xlabel={KL coefficient ($\beta$)},
    ylabel={Success Rate},
    xmin=-0.00, xmax=3.0,
    xtick={0,1,2,3},
    xticklabels={$0$,$0.03$,$0.3$,$3$},
    ymin=0.18, ymax=0.45,
    ytick={0.20,0.25,0.30,0.35,0.40,0.45},
    yticklabel style={/pgf/number format/fixed,
                      /pgf/number format/fixed zerofill,
                      /pgf/number format/precision=2},
    axis line style={black},
    tick style={black},
    tick label style={font=\small},
    label style={font=\small},
    grid=both,
    major grid style={black!25},
    minor grid style={black!35, dotted, line width=0.4pt},
    minor y tick num=1,
    minor x tick num=0,
    xminorgrids=false,
    legend style={
        draw=none,
        fill=none,
        font=\small,
        at={(0.02,0.02)},
        anchor=south west,
        legend cell align=left
    }
]

\addplot[draw=none,name path=rat-up,forget plot]
    table[x=idx, y expr=\thisrow{y}+\thisrow{err}] {\rationaledata};
\addplot[draw=none,name path=rat-lo,forget plot]
    table[x=idx, y expr=\thisrow{y}-\thisrow{err}] {\rationaledata};
\addplot[rationale,fill opacity=0.20,forget plot]
    fill between[of=rat-up and rat-lo];

\addplot[color=rationale, very thick, mark=*,
         mark options={fill=rationale,draw=rationale}, mark size=2pt]
    table[x=idx, y=y] {\rationaledata};
\addlegendentry{Rationale Mask}

\addplot[draw=none,name path=uni-up,forget plot]
    table[x=idx, y expr=\thisrow{y}+\thisrow{err}] {\uniformdata};
\addplot[draw=none,name path=uni-lo,forget plot]
    table[x=idx, y expr=\thisrow{y}-\thisrow{err}] {\uniformdata};
\addplot[uniform,fill opacity=0.18,forget plot]
    fill between[of=uni-up and uni-lo];

\addplot[color=uniform, very thick, mark=square*,
         mark options={fill=uniform,draw=uniform}, mark size=2pt]
    table[x=idx, y=y] {\uniformdata};
\addlegendentry{All-token Mask}

\addplot[black!60, dashed, thick, domain=0:3, samples=2] {0.362};
\addlegendentry{Without KL}

\end{axis}
\end{tikzpicture}
    \caption{Success rate of Qwen-3.5 4B on TimeWarp v6 as a function of the anchor strength $\beta$ with $\mathcal{L}_\text{WD}$, anchoring the rationale tokens or all tokens. The dashed line is without KL.}
    \label{fig:kl-success-rate}
\end{minipage}%
\vspace{-1em}
\end{figure*}

\subsection{Results}
\label{sec:results}

\paragraph{\name~outperforms every baseline on the seen distribution (RQ1).}
\name~improves over the strongest baseline by 10.7pp for Qwen~3.5~4B, 8.1pp for Qwen~3.5~9B, and 5.5pp for Gemma~4~E4B, on the seen TimeWarp version v6 (Tab.~\ref{tab:main_results}). Zero-shot students are near 0\% on every version, so the task is out of reach without training. Behavior cloning is competitive only when non-action tokens are included in the loss (BC-TW). No single alternative is reliably second best. m-GRPO is the strongest baseline on both Qwen models, but falls on Gemma. The other teacher-text baselines show the opposite pattern. Off-policy takeover is the strongest baseline on Gemma (34.3\%), but every teacher baseline trail m-GRPO on either the Qwen models. 

\paragraph{The gains hold on unseen versions and transfer to the live web (RQ2).}
On the unseen versions v1--v5, \name~beats the strongest baseline by 3.0--10.6pp. Version shift itself costs little once a student is trained, and the seen--unseen gap stays below 5pp. WebKD's gains hold on the live web, where Online-Mind2Web's numbers are low for all methods (below 8\%, Fig.~\ref{fig:om2Wresults}). \name~is the only method that clearly separates from the rest, roughly doubling the strongest baseline for the 9B student. The gains are mostly concentrated on the \texttt{easy} tasks, which have a shorter horizon.


\begin{table}[t]
    \centering

    \begin{minipage}[t]{0.61\textwidth}
        \vspace{0pt}
        \centering
        \small
        \setlength{\tabcolsep}{3.5pt}
        \renewcommand{\arraystretch}{1.10}

        \caption{
            \textbf{Edit--style ablation.}
            Teacher edits paired with in-context student style perform best. Qwen-3.5 4B baselines are
            trained using $\mathcal{L}_\text{WD}$, KL rationale anchor at $\beta = 0.03$.
        }
        \label{tab:edit-style-ablation}

        \begin{tabularx}{\linewidth}{
            c c >{\raggedright\arraybackslash}X c
        }
            \toprule
            $\mathbf{ES}$ & $\mathbf{SS}$
            & \textbf{Variant}
            & \textbf{S.R. $\uparrow$} \\
            \midrule

            T & T
            & Teacher takeover \textit{(off-policy)}
            & $17.48_{\pm 2.16}$ \\

            T & T
            & Teacher edit + teacher style
            & $33.01_{\pm 2.68}$ \\

            \rowcolor{cyan!06}
            T & S$_{\text{ic}}$
            & Teacher edit + student style \textit{(in-context)}
            & {$\mathbf{41.10_{\pm 2.80}}$} \\

            T & S$_{\text{inf}}$
            & Teacher edit + student re-style \textit{(2-pass)}
            & $25.89_{\pm 2.49}$ \\

            \midrule

            S/T & S/T
            & Student proposes, teacher judges/fixes
            & $27.83_{\pm 2.55}$ \\

            S & S
            & Student-only edit
            & $24.27_{\pm 2.44}$ \\
            \bottomrule
        \end{tabularx}

        \vspace{2pt}
        \raggedright
        \scriptsize
        $ES$: Edit Source; $SS$: Style Source.
        T = Teacher, S = Student.
    \end{minipage}
    \hfill
    \begin{minipage}[t]{0.37\textwidth}
        \vspace{0pt}
        \centering
        \small
        \setlength{\tabcolsep}{4pt}
        \renewcommand{\arraystretch}{1.10}

        \caption{
            \textbf{Loss ablation.}
            Token weightings on the filtered teacher rollouts, each with and without the
            KL rationale anchor at $\beta = 0.03$ using the Qwen-3.5 4B model.
        }
        \label{tab:loss-ablation}

        \begin{tabular}{lcc}
            \toprule
            \textbf{Loss}
            & \textbf{w/ KL}
            & \textbf{w/o KL} \\
            \midrule

            $\mathcal{L}_\text{CE}$
            & {$37.43_{\pm2.75}$}
            & {${36.25_{\pm2.73}}$} \\

            \rowcolor{cyan!6}
            $\mathcal{L}_\text{WD}$
            & {$\mathbf{41.10_{\pm2.80}}$}
            & {$\mathbf{38.51_{\pm2.77}}$} \\

            $\mathcal{L}_\text{GND}$
            & $35.92_{\pm2.73}$
            & $36.03_{\pm2.73}$ \\

            $\mathcal{L}_\text{CONS}$
            & $37.54_{\pm2.75}$
            & $35.81_{\pm2.73}$ \\

            $\mathcal{L}_\text{PPL}$
            & $37.54_{\pm2.75}$
            & $33.01_{\pm2.68}$ \\

            \bottomrule
        \end{tabular}
        \vspace{2pt}
    \end{minipage}
\vspace{-5pt}
\end{table}

\paragraph{The teacher should edit in the student's style, but minimally (RQ3).}
Following Tab.~\ref{tab:edit-style-ablation}, off-policy takeover underperforms as it drifts too far from the states the student has visited. Edits or suggestions written by the student are worse still, which we can attribute to the student lacking a reliable sense of its own correctness. Asking the student to restyle the teacher's edit on a second pass can also undermine the rationale. This is most evident on Qwen~3.5~9B, where the student re-style edit is a weaker baseline (Tab.\ref{tab:main_results}). Although vanilla teacher editing works, its outputs deviate more from the student's style and carry a larger perplexity premium (Fig.\ref{fig:perplexityPremium}). In-context student re-styling, as used by \name, stays closer to the student distribution and outperforms it by 8.1pp.

\paragraph{Loss should emphasize action informativeness, and anchor the rationale(RQ4).}
Tab.~\ref{tab:loss-ablation} compares token weightings of the loss, with and without the KL anchor on the rationale. $\mathcal{L}_\text{WD}$ with the kl anchor (Eq.\ref{eq:kl}) outperforms the other token-weighting schemes (\S\ref{subsec:lossAblationBaselines}) by 3.6pp. Sweeping the anchor strength shows that $\beta=0.03$ gives the highest success with rationale masking (Fig.~\ref{fig:kl-success-rate}).
\vspace{-1.5em}
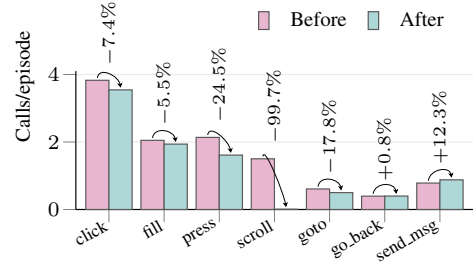
\begin{wrapfigure}{r}{0.45\textwidth}
  \centering
  \resizebox{\linewidth}{!}{\begin{tikzpicture}
\begin{axis}[
  width=0.45\columnwidth, height=4.5cm,
  font=\scriptsize,
  ybar=0pt, bar width=0.42,  
  ymin=0, ymax=7.3,          
  clip=false,                
  ylabel={Calls/episode},
  ytick={0,2,4},
  xtick={1,...,7},
  xticklabels={click,fill,press,scroll,goto,
               go\_back,send\_msg},
  x tick label style={font=\tiny,rotate=30,anchor=east},
  tick label style={inner sep=1pt},
  axis lines*=left,
  y axis line style={draw=none}, 
  tick align=outside,
  ymajorgrids, grid style={gray!15},
  enlarge x limits={abs=0.6}, 
  legend columns=2,
  legend style={
    draw=none, font=\scriptsize,
    at={(0.7,0.7)}, anchor=south,
    inner sep=0pt, /tikz/column sep=5pt
  },
  legend image code/.code={
    \draw[#1] (0pt,-2.8pt) rectangle (3.5pt,3.2pt);
  },
]
\addplot[
  fill=before, draw=black!65, thin
] coordinates {
  (1,3.828) (2,2.049) (3,2.136) (4,1.502)
  (5,0.608) (6,0.395) (7,0.783)
};
\addplot[fill=after,draw=black!65,thin] coordinates {
  (1,3.544) (2,1.937) (3,1.612) (4,0.005)
  (5,0.500) (6,0.398) (7,0.879)
};
\legend{Before,After}
\pgfplotsextra{
  \draw (rel axis cs:0,0) -- ({rel axis cs:0,0} |- {axis cs:1,4.1});
  \foreach \x/\before/\after/\change in {
    1/3.828/3.544/-7.4,
    2/2.049/1.937/-5.5,
    3/2.136/1.612/-24.5,
    4/1.502/0.005/-99.7,
    5/0.608/0.500/-17.8,
    6/0.395/0.398/+0.8,
    7/0.783/0.879/+12.3
  }{
    \draw[
      -{Stealth[length=1.5pt,width=1.6pt]},
      black, line width=0.3pt
    ]
      ([yshift=0.8pt]axis cs:\x-0.21,\before)
        .. controls +(0.6pt,4pt) and +(-0.6pt,4pt) ..
      ([yshift=0.8pt]axis cs:\x+0.21,\after);

    \pgfmathsetmacro{\topy}{max(\before,\after)}
    \node[font=\tiny,rotate=90,anchor=west,inner sep=0.5pt]
      at ([yshift=4.5pt]axis cs:\x,\topy)
      {$\change\%$};
  }
}
\end{axis}
\end{tikzpicture}}
  \caption{Change in action distribution before and after teacher distillation using \name for the Qwen-3.5 4B baseline.}
  \vspace{-2em}
\label{fig:actionDistributionChange}
\end{wrapfigure}
\paragraph{Discussion.} Behavior cloning for more epochs doesn't simply improve performance like \name~(\S\ref{subsec:longerBehaviorCloning}).
\textit{Teacher capability matters.} With the same prompts, states, and style exemplars, using Claude Opus~5 or Qwen-3.8 27B as the editor trains students to 33.3\% and 26.9\%, versus 41.1\% with GPT-5.5. As Qwen-3.8 is a smaller, less-performant model, the distilled student's performance is worse. 
\textit{Why can't students self-improve?} On average, self/student edit falls below BC~(TW) on every backbone (Tab.~\ref{tab:main_results}). This is consistent with the findings on RQ3, \textit{i.e.}, without an external signal, the student cannot reliably tell which of its own decisions were wrong.
\textit{How does \name~affect the behavior of the student?} It mostly prunes redundant actions, \textit{e.g}, scrolling nearly disappears ($-99.7\%$) and key presses drop by 24.5\% (Fig.~\ref{fig:actionDistributionChange}). \texttt{send\_msg} rises by 12.3\%, indicating higher task completion rate.
\textit{Where do the gains come from?} Our analysis reveals \name's strongest gains come from verifying the steps and answer {+14.6}pp and exiting loops {+9.4}pp. Detailed failure analysis (\S\ref{subsec:ErrorAnalysis}), and analysis of the improvements against behavior cloning (\S\ref{subsec:whereDogainsComeFrom}) have been provided in the Appendix.

\vspace{-5pt}
\section{Conclusion}
\vspace{-5pt}
Small, open-weight web agents are key to making web use automation widespread and accessible. Frontier agents have become highly capable, and distilling them into smaller models is a natural next step. Yet closed frontier models expose only their generated text, which rules out logit-based distillation. \name~addresses this by turning the teacher into an editor of the student's failures. It rewrites failed episodes in the student's own style and trains on them, using an objective that weights tokens based on action informativeness and a role-asymmetric KL anchor. Although \name~produced strong gains, the student still reaches only about half of the teacher's success rate, leaving substantial room for improvement. We hope this work spurs further research on black-box distillation and helps narrow the gap between closed frontier models and small open-weight agents.
\newpage
\clearpage

\bibliography{iclr2026_conference}
\bibliographystyle{iclr2026_conference}

\appendix

\section{Additional Background Details}
\label{sec:BackgroundDetails}

\subsection{Behavior Cloning}
\label{subsec:BehaviorCloning}
Given a corpus $\mathcal D$ of (context, target) pairs, behavior cloning is performed via supervised fine-tuning, which minimizes the teacher-forced cross-entropy \citep{ouyang2022training},
\begin{equation}
\mathcal L_{\mathrm{BC}}(\theta) \;=\;
-\;\mathbb{E}_{(x,y)\sim\mathcal D}\; \frac{1}{|y|}\sum_{i=1}^{|y|}\log \pi_\theta\!\left(y_i \mid x,\, y_{<i}\right).
\label{eq:sft}
\end{equation}
Behavior cloning is typically performed on the action tokens. However, the TimeWarp version \citep{ishmam2026timewarp} uses the full response, including the reasoning traces, and is the default type of {behavior cloning in our experiments}. We note that in Eq.~\ref{eq:sft}, every target token carries the same weight $1/|y|$, whether it {belongs to} the executed action or {to} filler thinking tokens. The state distribution is the teacher's, not the student's. At deployment, the policy conditions on its own generations and observations, and errors compound, which is the classic distribution-shift argument of imitation learning, under which a per-step error rate $\varepsilon$ can inflate to $O(\varepsilon T^2)$ cost over a $T$-step horizon unless supervision is collected on the
student's own state distribution \citep{ross2011dagger}.

\subsection{Knowledge Distillation}
\label{subsec:knowledgeDistillation}
Distillation replaces gold targets with supervision from a stronger teacher
\citep{hinton2015distilling}. For autoregressive models, the token-level form matches full
next-token distributions along a sequence,
\begin{equation}
\mathcal L_{\mathrm{KD}}(\theta) \;=\;
\mathbb{E}_{x}\,\mathbb{E}_{y}\,\sum_{i}\KL\!\big(p_{T}(\cdot \mid x, y_{<i})\,\big\Vert\,\pi_\theta(\cdot \mid x, y_{<i})\big),
\label{eq:kd}
\end{equation}
where the choice of the sequence $y$ (teacher-generated, gold, or student-generated) and of the divergence defines the family. Sequence-level KD \citep{kim2016sequence} sidesteps the inner distributions entirely and applies Eq.~\ref{eq:sft} to teacher-generated text. On-policy
distillation evaluates the divergence at prefixes the \emph{student} samples, correcting the
train--inference mismatch \citep{agarwal2024onpolicy}.

\subsection{Black-box Constraint}
\label{subsec:blackBoxConstraint}
Every token-level instantiation of Eq.~\ref{eq:kd} needs two things from the teacher: its full next-token distribution $p_T(\cdot \mid \cdot)$, and a shared tokenizer. The strongest available teachers for web agents are black-box models that usually provide neither. The distillation channel is
therefore \textit{only textual}: the teacher writes, and the student trains on the writing via Eq.~\ref{eq:sft}.
But text entangles two signals, (a) the teacher's \emph{decisions} (which element to click,
what to answer) and (b) its \emph{surface distribution} (style, phrasing, tone). Training a small
student on off-policy frontier text imports both at once, and the student pays a significant
\textit{perplexity premium} on the teacher's style (Fig.~\ref{fig:perplexityPremium}). 

\subsection{{Scratchpad Memory}}
\label{subsec:scratchpadMemory}
{The scratchpad belongs to the \emph{harness} state, not to the environment. It is the
sequence of memory components $(y^{\texttt{memory}}_1, \ldots, y^{\texttt{memory}}_{t-1})$ the
agent has emitted so far, carried inside the response history $y_{1:t-1}$ that the harness
places in every context $h_t$ (\S\ref{subsec:background}). The agent interacts with it only
through the memory tokens. It \emph{writes} by emitting $y^{\texttt{memory}}_t$ and
\emph{reads} only what the harness surfaces back at later steps.} Under truncated history, a fact from a page the agent has since left survives only if it was written into a rationale component, most durably into the scratchpad memory.

\subsection{{Extended Problem Formulation}}
\label{app:pomdp}

Details deferred from the Background's POMDP formulation
$\langle \mathcal S, \mathcal A, \mathcal O, \mathbb T, \mathbb T_0, \mathbb O, R \rangle$
\citep{kaelbling1998planning}.

\paragraph{Observation modalities.}
The observation function $\mathbb O : \mathcal S \to \mathcal O$ may render the browser state as
raw HTML, a set-of-marks screenshot \citep{yang2023setofmark}, or an accessibility tree. We use
the AXTree throughout, a textual serialization of the rendered page, annotated with the
element identifiers the action space refers to.

\paragraph{Action space.}
$\mathcal A$ is a subset of BrowserGym's browser-level action space
\citep{dechezelles2025browsergym}: page interactions (e.g.\ \texttt{click}, \texttt{fill}),
navigation (e.g.\ \texttt{goto}), tab operations (e.g.\ \texttt{tab\_focus}), and user
interaction (\texttt{send\_msg\_to\_user}, which is also how the agent submits its answer). The generated action is deterministically parsed from the \texttt{action} component of the response and
passed to the browser for execution. The transition
$\mathbb T : \mathcal S \times \mathcal A \to \mathcal S$ is treated as deterministic.

\paragraph{Initial states and episodes.}
$\mathbb T_0$ draws the initial state uniformly over the benchmark's tasks. An episode ends when
the agent messages the user or exhausts its step budget, at which point the LLM-judge evaluator
\citep{zheng2023judging} assigns the terminal $R \in \{0,1\}$ against the task specification.

\paragraph{Environment versions.}
The environment comes in multiple versions $\{\mathcal E_1, \ldots, \mathcal E_n\}$, the same
sites captured at different points in time, each defining its own instance of the POMDP
tuple. Training and evaluation fix one version, except when cross-version transfer is itself
under test.

\section{Methodology Details}
\label{sec:methodologyDetails}

\subsection{Teacher Details}
\label{app:teacher}

\paragraph{Teacher-as-agent.}
The teacher is prompted with the \emph{same} rendered system and user prompt the student saw:
identical goal, AXTree serialization, history format, and action-space description. This keeps the
teacher's output in-distribution as an SFT target rather than free-form advice, and makes the edit
a drop-in replacement for the student's own response. The teacher receives no privileged input in
our method: no gold walkthrough, no hints.

\paragraph{Localization.}
{The teacher makes one call per failed episode over the full logged trajectory.} The instruction reads:
\begin{quote}\small\ttfamily
The agent's trajectory above FAILED. Identify the FIRST erroneous step -- the earliest step whose
action set the trajectory on the path to failure. Reply with ONLY the integer step index
(0-based). Nothing else.
\end{quote}

\paragraph{Section contract.}
Every edit query ends with a format contract requiring exactly the five tags \texttt{<think>},
\texttt{<plan>}, \texttt{<step>}, \texttt{<memory>}, \texttt{<action>}: \textit{think-aloud} reasoning grounded only in
what is visible in the observation, the running multi-step plan, a single integer in
\texttt{<step>} naming the plan step the action belongs to, facts worth carrying forward in
\texttt{<memory>}, and exactly one executable action. The contract exists because the student is
trained and evaluated with the thinking, planning, and memory components enabled. A teacher that
omits one of these components would silently teach the student to drop the field.

\paragraph{Style block.}
{The $K{=}3$ exemplars are appended \emph{after} the contract, at the very end of the query.
The contract is the instruction whose loss is most expensive (a missing tag costs the whole
step), so it must not be pushed out of the recency window, whereas a style hint that is only
partly followed still leaves a usable record. Exemplar selection is deterministic given the task.
Solved episodes are preferred, the task being edited is excluded, and at most one exemplar is
drawn per source task. Exemplars carry the rationale components and the step index but never
\texttt{<action>}.}

\paragraph{Correction prompt.}
The correction query is assembled as the student's rendered context,
then the section contract, then the style block. The style block opens with the following
instruction:
\begin{quote}\small\ttfamily
[STYLE -- write this step in the student agent's own style]\\
The five sections you emit become training data for the agent whose steps appear above. Match ITS
writing style: same register, same typical length, same amount of detail, the same way of
phrasing a plan and of noting a memory. Below are verbatim examples of how THIS agent writes.\\[4pt]
Copy the STYLE only. Do NOT copy their content, their decisions, or their conclusions -- they are
unrelated steps from other tasks. Your decision for this step must remain exactly the one you
would have made without these examples; only the wording changes.
\end{quote}
It is followed by the $K$ exemplars, each delimited by
\texttt{how the student writes, example $n$} and \texttt{end example $n$}. The correction call is made once per step $k \ge t'$, teacher-forced on the
student's original prefix. 


\begin{figure}[t]
\begin{tcolorbox}[colback=gray!08, colframe=gray!8, boxrule=0pt, arc=5pt,
                  left=2pt, right=4pt, top=6pt, bottom=4pt]
\begin{algorithmic}[1]
\REQUIRE base model $\pi_{\theta_0}$, black-box teacher $\pi_T$ (text only), training tasks, exemplar count $K$, grounding threshold $\rho_{\min}$, weight floor $\phi$, dead zone $\tau_{\mathrm{dz}}$, anchor strength $\beta$
\ENSURE $\pi_{\theta_{\mathrm{WD}}}$
\STATE \textbf{P1 (seed):} $\pi_{\theta_\text{BC}} \gets$ fine-tune $\pi_{\theta_0}$ on successful teacher rollouts $\mathcal D_\text{T}$ with $\mathcal L_{\mathrm{BC}}$ (Eq.~\ref{eq:sft}); freeze
\STATE \textbf{P2 (rollout):} run $\pi_{\theta_\text{BC}}$ on every training task; split episodes into failures $\mathcal D_{f(\text{BC})}$ and style pool $\mathcal D_{s(\text{BC})}$
\STATE $\mathcal D \gets \emptyset$
\FORALL{$\tau = (h_1, y_1, \ldots, h_T, y_T) \in \mathcal D_{f(\text{BC})}$}
    \STATE \textbf{P3 (localize):} $t' \gets \mathrm{Loc}(\tau)$, the earliest erroneous step (estimate of $t^\star$, Eq.~\ref{eq:tstar})
    \STATE \textbf{Replay:} $\mathcal D \gets \mathcal D \cup \{(h_k, y_k)\}$ for $k = 1, \ldots, t'-1$
    \STATE \textbf{P4 (edit):} $y^*_k \sim \pi_T(\cdot \mid h_k,\, \mathcal D_{s(\text{BC})})$ with $K$ style exemplars, unexecuted (Eq.~\ref{eq:edit}), for $k = t', \ldots, T$
    \STATE \textbf{Filter:} $\mathcal D \gets \mathcal D \cup \{(h_k, y^*_k) : \rho_k \ge \rho_{\min}\}$
\ENDFOR
\STATE \textbf{Weights:} for each edited $(h_k, y^*_k) \in \mathcal D$, score segments by $\Delta_s$ under $\pi_{\theta_\text{BC}}$ (Eq.~\ref{eq:actw}), rank-map to $v_s \in [\phi, 1]$ (Eq.~\ref{eq:actw-map}), spread to token weights $w_i$ (Eq.~\ref{eq:chartok}); pin $\langle\mathrm{A}\rangle, \langle\mathrm{S}\rangle$ tokens at $w_i = 1$; replayed records keep uniform weights
\STATE \textbf{Mask:} $\gamma_i \gets 1$ on $\langle\mathrm{T}\rangle, \langle\mathrm{P}\rangle, \langle\mathrm{M}\rangle$ tokens; $\gamma_i \gets 10^{-6}$ on $\langle\mathrm{A}\rangle, \langle\mathrm{S}\rangle$ tokens (Appendix~\ref{app:kl-impl})
\STATE \textbf{Train:} $\pi_{\theta_{\mathrm{WD}}} \gets \arg\min_\theta \mathcal L_{\mathrm{WD}}$ (Eq.~\ref{eq:kl}) over $\mathcal D$, initialized at $\pi_{\theta_\text{BC}}$ with anchor $\pi_{\text{ref}} = \pi_{\theta_\text{BC}}$
\RETURN $\pi_{\theta_{\mathrm{WD}}}$
\end{algorithmic}
\vspace{2pt}\hrule\vspace{6pt}
\captionof{algorithm}{{\name: black-box distillation through grounded teacher edits}}
\label{alg:webkd}
\end{tcolorbox}
\end{figure}

\subsection{P3 Failure Localization Details}
\label{subsec:p3failureLocalizationDetails}

\begin{definition}[Recoverability]
\label{def:recover}
For a trajectory $\tau$, let $\nu_t \in \{0,1\}$ indicate that, from the underlying browser state reached after the first $t$ actions of $\tau$, \emph{some} action sequence within the remaining step budget $T-t$ ends with $R=1$.
\end{definition}

\begin{proposition}[Every failed web trajectory has a unique first fatal step]
\label{prop:tstar}
Assume the harness of \S\ref{subsec:background}: a finite step budget $T$, deterministic transitions and parsing, a terminal binary reward, and a task solvable from its initial state ($\nu_0 = 1$). Then along any failed trajectory $\nu_t$ is non-increasing {and equals $0$ at the
episode's end}, so
\begin{equation}
t^\star \;=\; \min\{t \ge 1 : \nu_t = 0\}
\label{eq:tstar}
\end{equation}
exists and is unique: every step before $t^\star$ leaves the task solvable, every step from $t^\star$ on leaves it unsolvable, and at step $t^\star$ a recoverability-preserving action existed but was not taken. (Proof and discussion in \S~\ref{app:tstar}.)
\end{proposition}

The two ways $t'$ can miss $t^\star$ have very different costs. If the teacher localizes too early ($t' < t^\star$), it rewrites steps from which the task was still solvable. The state at $t'$ is still recoverable, so the only cost is extra edits to the corpus. If it localizes too late ($t' > t^\star$), the replayed prefix includes, and so endorses, the fatal steps, and by Proposition~\ref{prop:tstar} no continuation from $h_{t'}$ can succeed. This asymmetry is why the localization prompt asks for the \emph{earliest} erroneous step (\S\ref{app:teacher}), and it shows up again as the one-sided error $\varepsilon_{\mathrm{loc}}$ in the guarantee of \S\ref{sec:bound}.

\paragraph{Localization Accuracy.} We sample $200$ failed episodes from the rollout pools of the three students. Two human annotators, blind to the teacher's prediction and to each other, independently mark the earliest step after which success within the remaining budget became impossible, following the same recoverability instruction given to the teacher. Under the strict budget reading, a step counts as unrecoverable only when the remaining budget is insufficient to complete the task. Here, the teacher is never late. It is early on $93\%$ of episodes and exact on the remaining $7\%$. Under the loop-onset reading, the failure point is the step at which the agent enters a loop that it never escapes. Here the teacher is early on about two-thirds of episodes, exact on about a quarter, and late on roughly $10\%$. Under both readings, its errors fall mostly on the early side, which is the benign direction. {\color{black}The budget reading is the $t^\star$ of Proposition~\ref{prop:tstar}, so its late rate is the $\varepsilon_{\mathrm{loc}}$ of Theorem~\ref{thm:bound}.} 

\subsection{P4 Teacher Edit Details}
\label{subsec:teacherEditDetails}

\emph{(i) Style conditioning.} The teacher's query ends with a style block containing $K{=}3$
verbatim exemplars of the student's own writing. The exemplars are mined from the style pool
$\mathcal D_{s(\text{BC})}$, excluding the task being edited, and are selected deterministically
per task, with at most one exemplar per source task. The block opens with an explicit instruction
to match this agent's register, typical length, and level of detail, and to copy only the style,
keeping the decision ``exactly the one you would have made without these examples, but only the
wording changes'' (Appendix~\ref{app:teacher}). Exemplars include the student's rationale
components but \emph{never} its $\langle\mathrm{A}\rangle$ component: an action is a decision, and
the style block should transfer the student's style, not its decisions. This directly targets the
perplexity premium of \S\ref{subsec:perplexityPremium}. Apart from the correction itself, the corrected
trajectory should read as low-surprise text to the student.

\emph{(ii) Edits are never executed.} The environment always steps on the student's
\emph{original} action. An edit at step $k$ therefore leaves the observation at step $k{+}1$
unchanged, corrections never compound, and every $h_k$ in the corpus is a state the student
actually reached. Each teacher call conditions on the student's original prefix, never on an earlier edit. This mirrors the central idea of on-policy
distillation, where the supervising distribution is evaluated on prefixes that the student itself produced \citep{agarwal2024onpolicy,zhao2026opsd}.

\subsection{\name~Loss Details}
\label{subsec:WebKDLossDetails}

\begin{remark}[$\Delta_s$ {\color{black}as} a conditional pointwise mutual information]
\label{rem:pmi}
Writing $c = (h,\, {\color{black}y^*_{<a}} \setminus s)$ for the context with $s$ deleted,
\begin{equation}
\Delta_s \;=\; \log \frac{P_{\pi_{\theta_\text{BC}}}\!\left(\langle\mathrm{A}^*\rangle \mid s,\, c\right)}{P_{\pi_{\theta_\text{BC}}}\!\left(\langle\mathrm{A}^*\rangle \mid c\right)}
\;=\; \mathrm{PMI}_{\pi_{\theta_\text{BC}}}\!\left(\langle\mathrm{A}^*\rangle ;\, s \mid c\right), 
\label{eq:pmi}
\end{equation}
{\color{black}which reads as} the pointwise mutual information, under the student's own distribution, between the segment and
the decision it precedes. {\color{black}The first equality is exact. The second takes the student's probability given the shortened prefix as the reference $P(\langle\mathrm{A}^*\rangle \mid c)$, and deleting $s$ is not marginalizing over it, so it is an interpretation of the log-ratio rather than an identity.} {Negative $\Delta_s$ is meaningful as it marks
segments whose presence makes the student \emph{less} certain of the correct decision, {which is why such segments stay in the ranking and are never masked}.}
\end{remark}

{We use ranks rather than raw magnitudes by default. Because
$\log P(\langle\mathrm{A}^*\rangle)$ varies by two orders of magnitude across steps, a
magnitude-based weighting would implicitly weight segments by action length. Ranking within
each step removes this dependence on scale. 
Span
weights $v_s$ become token weights $w_i \in [\phi, 1]$ by character-coverage average, so a token
straddling a segment boundary receives the area-weighted blend (Eq.~\ref{eq:chartok},
Appendix~\ref{app:actw}).}

\paragraph{{Scoring policy and the complete map.}}
{Segments are sentences or lines of the $\langle\mathrm{T}\rangle$, $\langle\mathrm{P}\rangle$,
and $\langle\mathrm{M}\rangle$ components. $\langle\mathrm{S}\rangle$ and $\langle\mathrm{A}\rangle$
are never segmented, and the merge rules and caps are given in Appendix~\ref{app:actw}. Both
terms of Eq.~\ref{eq:actw} are scored once, offline, under the \emph{frozen seed checkpoint}
$\pi_{\theta_\text{BC}}$, the same policy that provides the initialization and, in
\S\ref{sec:kl}, the anchor, so the weights describe the student the training starts from. We write
$r_s \in [0,1]$ for the normalized average rank of $\Delta_s$ within its step (ties share their
mean rank), $\sigma = \max_s \Delta_s - \min_s \Delta_s$ for the within-step spread in nats, and
$\tau_{\mathrm{dz}} > 0$ for the dead-zone width. Then}
\begin{equation}
v_s \;=\; \mathrm{clip}_{[\phi,\,1]}\!\Big(\phi + (1-\phi)\Big(\tfrac{1}{2} + \big(r_s - \tfrac{1}{2}\big)\min\!\big(1,\; \sigma/\tau_{\mathrm{dz}}\big)\Big)\Big),
\label{eq:actw-map}
\end{equation}
{where $\tau_{\mathrm{dz}} = 0.05$~nats.} {So a step whose segments are statistically indistinguishable ($\sigma \ll
\tau_{\mathrm{dz}}$) degrades to the midpoint weight $(1+\phi)/2$ on every segment {\color{black}(exactly so at $\sigma = 0$; within $(1-\phi)\sigma/(2\tau_{\mathrm{dz}})$ of it otherwise)}, which is the mean {\color{black}span weight of every} ranked
steps, since average ranks normalized to $[0,1]$ have mean $1/2$. A single-segment step has $\sigma = 0$ and lands at
the same midpoint.} {The imitation term of the objective is a \emph{weighted mean} of per-token cross-entropies,}
\begin{equation}
\mathcal{L}_{\text{SW}} \;=\; \frac{\sum_{i \in T} w_i\, \mathrm{CE}_i}{\sum_{i \in T} w_i},
\label{eq:sw}
\end{equation}
{where $T$ indexes the supervised response tokens and
$\mathrm{CE}_i = -\log \pi_\theta(y^*_i \mid h, y^*_{<i})$ is the cross-entropy of corpus token
$i$. The denominator is the \emph{weight sum}, never the token count. This matters because it
makes $w_i \equiv c$ reproduce standard SFT {\color{black}(the per-token mean over the same tokens)} exactly for any $c>0$, so a scheme whose weights
average $\bar w$ does not silently train at an effective learning rate of $\bar w \cdot \eta$,
for learning rate $\eta$.}


\begin{remark}[The update resamples tokens rather than rescaling the loss]
\label{rem:resample}
Since $q_w(i) = w_i / \sum_j w_j$ is a probability distribution over supervised positions,
$\mathcal L_{\mathrm{SW}} = \mathbb E_{i \sim q_w}\!\left[\mathrm{CE}_i\right]${, so gradient descent
on Eq.~\ref{eq:sw} is gradient descent on ordinary SFT over a corpus in which tokens are
resampled in proportion to their weights.}
\end{remark}
\subsection{{Action-Informative Weighting: Constants and Scoring Interface}}
\label{app:actw}

\paragraph{Segmentation.}
Boundaries are newlines, or whitespace following sentence-final punctuation
(\texttt{. ! ? ; :}). Each separator stays with the preceding segment, so segments exactly
partition each component's inner span. Segments shorter than 24 non-whitespace characters are
merged into the previous segment (into the next, if first). A step is capped at 24 segments by
repeatedly merging the shortest adjacent pair. Merges never cross a component boundary.

\paragraph{Scoring interface.}
Both terms of Eq.~\ref{eq:actw} are computed by teacher-forced echo-scoring against a served copy
of the seed checkpoint: the full (context, target) string is submitted with the prompt echoed and
per-token log-probabilities returned, at temperature zero and with no new tokens generated. The
log-probability of $\langle\mathrm{A}^*\rangle$ is the sum over its token positions. The ablated variant deletes the
segment's characters (including its trailing separator) and re-tokenizes. Numbered plan lines are
deliberately \emph{not} renumbered after a deletion. Renumbering would perturb segments other than
the ablated one and break the leave-one-out contract. The chat template's pre-opened
\texttt{<think>} tag is stripped before the target is appended, so the scored string matches the
training-time rendering.

\paragraph{Characters to tokens.}
With $\mathrm{chars}(i)$ the character span of token $i$ under the tokenizer's offset mapping, the
token weight is the coverage-weighted average
\begin{equation}
w_i \;=\; \frac{\sum_{s} \left|\mathrm{chars}(i) \cap s\right| \cdot v_s \;+\; \big|\mathrm{chars}(i) \setminus \bigcup_s s\big| \cdot 1}{\left|\mathrm{chars}(i)\right|}\,,
\label{eq:chartok}
\end{equation}
clamped to $[\phi, 1]$. Characters outside every rationale segment (tags, separators, and the
components pinned at weight 1) contribute weight 1, zero-width special tokens take weight 1,
and the response envelope (the template's closing tokens) is never down-weighted.

\subsection{{Anchor Implementation}}
\label{app:kl-impl}

\paragraph{Estimator.}
$\mathrm{kl}_i$ in Eq.~\ref{eq:kl} is the ``$k_3$'' estimator \citep{schulman2020kl}: it is
non-negative and low-variance, and unbiased for $\KL(\pi_\theta \Vert \pi_{\text{ref}})$ under
sampling from $\pi_\theta$. The log-ratio $d_i$ is clamped to $[-20, 8]$. {The asymmetry is
deliberate. $d_i \approx 0$ at initialization and drifts \emph{negative} as the student fits the
corpus, where $k_3$ grows only linearly and is harmless.} The positive tail is the direction in
which $e^{d}$ can overflow 16-bit precision, and it is not the direction the penalty exists to
control.


\paragraph{Reference pass.}
$\pi_{\text{ref}}$ is the frozen seed checkpoint, held in 16-bit precision and evaluated in a
second, gradient-free forward pass per micro-batch. Per-token log-probabilities are gathered with
a fused cross-entropy rather than a full log-softmax, avoiding a $|y| \times |\mathcal V|$
32-bit intermediate. The anchor's marginal memory cost is nonetheless substantial, an
\emph{unsharded} 16-bit copy of the reference policy on every rank (measured $\approx$27\,GB at
9B scale) on top of the loss computation's 32-bit logit upcast, which is why training runs
single-rank with ZeRO-3 parameter partitioning and CPU-offloaded optimizer state
(Appendix~\ref{app:training}).

\paragraph{Why normalize by $\sum_i \gamma_i$.}
In our corpus, the rationale mask anchors roughly $88\%$ of supervised characters, while an action
mask would anchor roughly $9\%$. Normalized by token count instead, the same $\beta$ would be an
order of magnitude weaker under the action mask because actions are short.

\section{Additional Experimental Details}
\label{sec:experimentalDetails}
\subsection{Benchmark Baselines}
\label{subsec:benchmarkBaselines}

{We describe each baseline of Tab.~\ref{tab:main_results} and Fig.~\ref{fig:om2Wresults} below.}

\begin{enumerate}[leftmargin=20pt]
    \item \textbf{Zero-Shot:} {The base model without fine-tuning.}
    \item \textbf{Behavior Cloning (BC):} {The base model is} trained using supervised fine-tuning (Eq.~\ref{eq:sft}) on the teacher rollouts containing {the} observation history, prompts, and actions only.
    \item \textbf{Behavior Cloning (TW):} The TimeWarp version of behavior cloning \citep{ishmam2026timewarp}. {It is identical to vanilla BC, except that it also includes} the thinking, planning, step, and memory tokens during behavior cloning.
    \item \textbf{m-GRPO}: Multi-turn Group Relative Policy Optimization (m-GRPO) is an outcome-based RL baseline, which uses the trajectory's reward for credit assignment \citep{wei2025webagentr1}.
    \item \textbf{Self/Student Edit}: {An edit-style baseline in which the student, instead of an external teacher model, edits its own failed trajectories.}
    \item \textbf{Off-policy Takeover:} An edit-style baseline where, instead of the teacher editing the student's failed trajectories, the teacher {takes} over from the point of failure and {completes} the trajectory. The student is then trained \textit{off-policy} on the edited trajectories. This is similar to on-policy expert correction \citep{lauffer2025oec}.
    \item \textbf{Student Re-style Edit:} {A two-pass edit, in which} the teacher first edits the student's failed trajectories{, and the student is then prompted} to rewrite the edits in its own style. The student is then trained on-policy on the re-styled edits.
\end{enumerate}

\begin{table}[ht]
\caption{Training hyperparameters for the combined objective.}
\label{tab:hparams}
\begin{center}
\begin{tabular}{ll}
\toprule
\bf Hyperparameter & \bf Value \\
\midrule
fine-tuning & full-parameter, from the seed checkpoint \\
learning rate & $1\times10^{-5}$, cosine decay, warmup ratio $0.1$ \\
epochs & $2$ \\
context cutoff & $20{,}480$ tokens \\
batch & $1$ per device $\times$ $16$-step gradient accumulation (effective $16$) \\
precision & bf16, gradient checkpointing \\
partitioning & ZeRO-3 with CPU offload \\
CE weight floor $\phi$ & $0.1$ \\
dead zone $\tau_{\mathrm{dz}}$ & $0.05$ nats \\
anchor strength $\beta$ & $0.03$ (per anchored token) \\
anchor mask $\gamma$ & $1$ on \texttt{think}/\texttt{plan}/{\texttt{memory}}, $10^{-6}$ elsewhere \\
\bottomrule
\end{tabular}
\end{center}
\end{table}

\subsection{Loss Ablation Baselines}
\label{subsec:lossAblationBaselines}

Each variant keeps the weighted objective of Eq.~\ref{eq:sw} and changes only the token weights
$w_i \in [\phi, 1]$. The $\langle\mathrm{A}\rangle$ and $\langle\mathrm{S}\rangle$ tokens stay at
$w_i = 1$ in every variant. Span weights $v_s$ become token weights $w_i$ by the same
character-coverage average as in \name. The ``w/ KL'' column of Table~\ref{tab:loss-ablation} adds
the anchor of Eq.~\ref{eq:kl} unchanged.
\begin{enumerate}[leftmargin=20pt]
    \item $\mathcal{L}_\text{CE}$ (uniform). Standard SFT cross-entropy with equal weight on every
    response token, the control for all weighted variants.
    \item $\mathcal{L}_\text{GND}$ (grounding). A soft, within-step version of the grounding filter
    of Algorithm~\ref{alg:webkd}, which floors any rationale span that makes a claim not supported
    by the goal or by the AXTrees observed up to step $k$. The claims $\mathcal{C}(s)$ of a span are
    its quoted strings, numbers, element identifiers, and capitalized entities, and a claim is
    grounded if its case- and whitespace-folded form occurs in the goal or in one of those AXTrees,
    \begin{equation}
    v_s \;=\;
    \begin{cases}
        \phi & \text{if some } c \in \mathcal{C}(s) \text{ is not grounded,}\\
        1    & \text{otherwise.}
    \end{cases}
    \label{eq:gnd}
    \end{equation}
    \item $\mathcal{L}_\text{CONS}$ (teacher self-consistency). A span is weighted by how much of it
    the teacher reproduces when it re-edits the same step $k = 3$ times, net of the overlap expected
    by chance,
    \begin{equation}
    v_s \;=\; \phi + (1-\phi)\,
    \mathrm{clip}_{[0,1]}\!\left(\frac{c_\text{same}(s) - c_\text{null}(s)}{1 - c_\text{null}(s)}\right),
    \label{eq:cons}
    \end{equation}
    where $c_\text{same}(s)$ is the IDF-weighted containment of $s$ in the teacher's resampled edits
    of the same step and $c_\text{null}(s)$ is the same statistic against resampled edits of steps
    from other tasks.
    \item $\mathcal{L}_\text{PPL}$ (student perplexity). Each token is weighted by how likely the
    seed student already finds it, so the high-perplexity tokens of the teacher's edit are floored,
    \begin{equation}
    w_i \;=\; \phi + (1-\phi)\, r_i,
    \label{eq:ppl-weight}
    \end{equation}
    where $r_i \in [0,1]$ is the normalized average rank of
    $\log \pi_{\theta_\text{BC}}(y^*_i \mid \cdot)$ among the rationale tokens of its step.
\end{enumerate}

\subsection{{Training Configuration}}
\label{app:training}

\paragraph{Corpus.} The corpus records are step-level: one (context, response) pair per step, the context being the full rendered agent state (goal, AXTree, history) and the response the five-component output.
On the corpus used throughout, 15 failed training episodes yield 432 step records. The teacher
edits 403 of them (the remaining 29 precede $t'$ and are replayed), {the grounding filter of Algorithm~\ref{alg:webkd}} removes 40 edited
steps ($9.9\%$), and 392 records survive (363 edited, 29 replayed). The edited targets total
{$\approx$64k response tokens (16.6k \texttt{think}, 19.3k \texttt{plan}, 8.7k \texttt{memory}, 5.2k
\texttt{action}, 13.8k structural).} Rollouts (P2) use temperature $0.01$ and a 30-step budget,
with the thinking, plan, and memory components enabled and no hints, the same interface as
evaluation. Training is full-parameter fine-tuning from the seed checkpoint with the hyperparameters of
Table~\ref{tab:hparams}. 


\paragraph{Compute.}
Each fine-tune runs on a single H200 GPU with ZeRO-3 parameter partitioning and CPU-offloaded
optimizer state. Corpus generation and evaluation parallelize over a farm of
single-GPU workers, one containerized browser environment per GPU, and all teacher and judge
calls go through hosted APIs.

\paragraph{Cost and latency.} Most of the teacher cost is in P1, which makes
$4{,}448$ calls and uses $21.8$\,M prompt tokens to produce the demonstrations. The correction
stage is much cheaper. P3 makes one call per failed episode ($15$ calls, $0.47$\,M tokens), and P4
makes one call per edited step ($403$ calls, $4.0$\,M tokens, $6\%$ of which is the style block).
Correction is therefore $17\%$ of the pipeline's $26.3$\,M prompt tokens, paid once, or about the
cost of $135$ teacher episodes at the teacher's in-harness rate of $33$k prompt tokens per episode.
These counts are lower bounds, since the prompts of $13\%$ of the P1 conversations were not kept
and hidden reasoning tokens were not logged ($0.78$\,M visible completion tokens). A fine-tune
takes $1.0$ GPU-hour on one H200, and evaluating the $103$ v6 tasks takes $20$ minutes on one
$48$\,GB L40S. Measured inside the harness, the trained 4B student takes a median of $6.10$\,s per
step ($38.8$\,s per episode), against $9.05$\,s per step for the seed and $12.80$\,s per step
($68.1$\,s per episode) for the teacher acting as the agent.

{
\section{{Additional Experimental Results}}
\label{app:results}

\subsection{Can we simply train behavior cloning for longer?}
\label{subsec:longerBehaviorCloning}

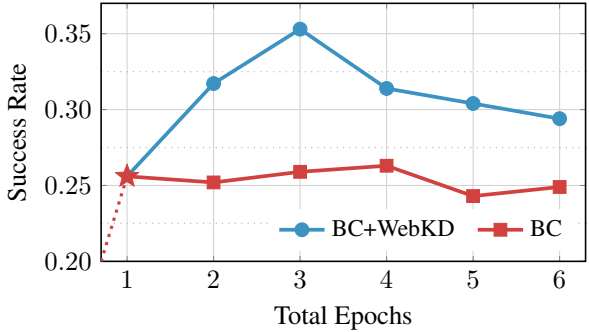
\begin{wrapfigure}{r}{0.575\textwidth}
  \centering
  \resizebox{\linewidth}{!}{\definecolor{cblue}{HTML}{3C93C2}
\definecolor{cred}{HTML}{D1423F}

\begin{tikzpicture}
\begin{axis}[
    width=8cm, height=5cm,
    xlabel={Total Epochs},
    ylabel={Success Rate},
    xmin=0.7, xmax=6.3,
    ymin=0.20, ymax=0.37,
    xtick={1,2,3,4,5,6},
    ytick={0.20,0.25,0.30,0.35},
    minor y tick num=1,
    yticklabel style={/pgf/number format/fixed,
                      /pgf/number format/fixed zerofill,
                      /pgf/number format/precision=2},
    xmajorgrids, ymajorgrids, yminorgrids,
    major grid style={draw=gray!35, line width=0.4pt},
    minor grid style={draw=gray!55, dotted, line width=0.5pt},
    tick align=inside,
    major tick length=2pt,
    minor tick length=0pt,
    axis line style={line width=0.6pt},
    every axis plot/.append style={line width=1.4pt, mark options={line width=0.4pt}},
    legend style={at={(0.98,0.04)}, anchor=south east,
                  draw=none, fill=white,
                  font=\small, row sep=-1pt,
                  /tikz/every even column/.append style={column sep=6pt}},
    legend columns=2,
    legend cell align=left,
]

\draw[cred, dotted, line width=1.2pt] (rel axis cs:0,0) -- (axis cs:1,0.256);

\addplot[color=cblue, mark=*, mark size=2.6pt, mark phase=2] coordinates {
    (1, 0.256)
    (2, 0.317152104)
    (3, 0.353)
    (4, 0.3139158576)
    (5, 0.304)
    (6, 0.294)
};
\addlegendentry{BC+WebKD}

\addplot[color=cred, mark=square*, mark size=2.4pt, mark phase=2] coordinates {
    (1, 0.256)
    (2, 0.252)
    (3, 0.259)
    (4, 0.263)
    (5, 0.243)
    (6, 0.249)
};
\addlegendentry{BC}

\node[star, star points=5, star point ratio=2.3,
      fill=cred, draw=cred, line width=0.4pt,
      inner sep=0pt, minimum size=9pt]
    at (axis cs:1,0.256) {};

\end{axis}
\end{tikzpicture}}
  \caption{Behavior cloning over more epochs doesn't improve the success rate, as it flattens at $\sim25\%$, whereas \name~continues to improve and peaks at the 3rd total epoch for the Qwen-3.5 9B model. \name~is trained from the first epoch of the BC seed.}
  \label{fig:trainingBClonger}
  \vspace{-20pt}
\end{wrapfigure}
We fine-tune Qwen-3.5 9B using the TimeWarp version of Behavior Cloning (BC) for longer epochs, and compare it against the standard setting of \name~where we fine-tune from the BC seed. Fig. \ref{fig:trainingBClonger} shows the success rate of these two baselines for the Qwen-3.5 9B model. BC alone plateaus early, remaining between 0.24 and 0.26 regardless of the training budget. This suggests that additional epochs on the same training data alone provide little benefit. In contrast, adding \name~results in consistent gains at every budget beyond the one-epoch BC seed. The best checkpoint is when \name~is trained for two additional epochs.


\vspace{0.25cm}
\subsection{Error Analysis}
\label{subsec:ErrorAnalysis}

\begin{figure}[ht]
    \centering
\includegraphics[width=0.85\linewidth]{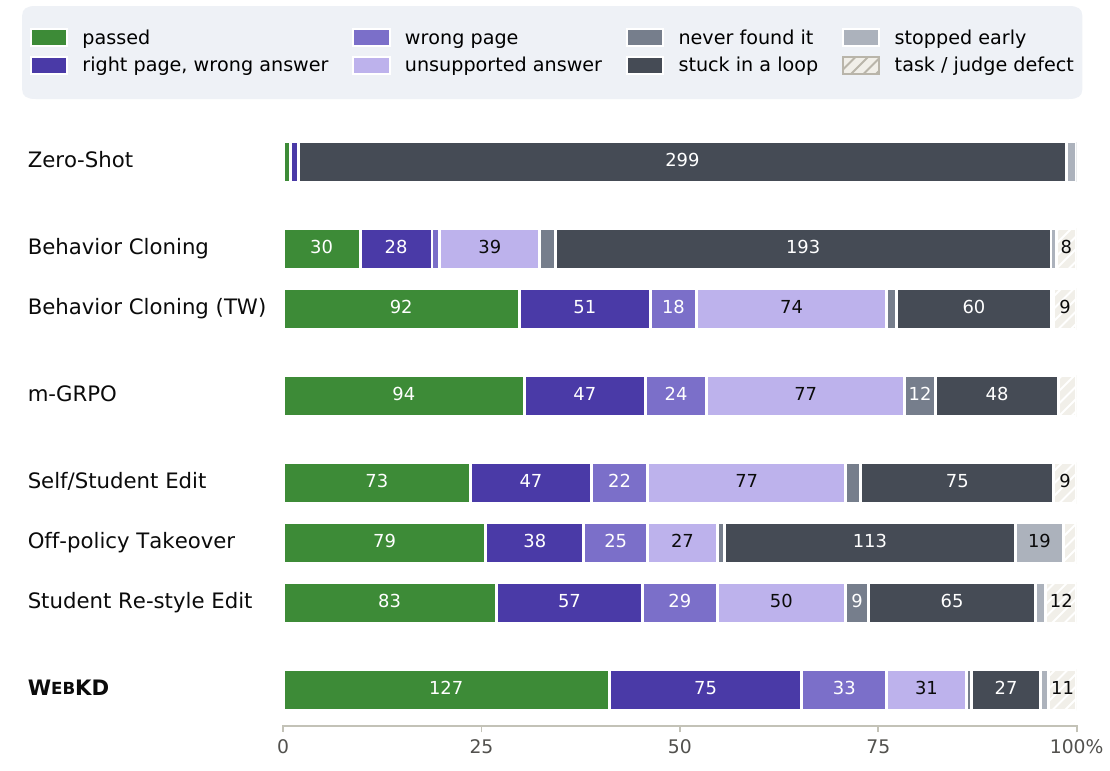}
    \caption{\textbf{Failure mode analysis of Qwen3.5-4B.} Each bar assigns every evaluation trajectory of a method to exactly one outcome for the $103\times3=309$ tasks.}
    \label{fig:failureModeErrorAnalysis}
\end{figure}

Fig.~\ref{fig:failureModeErrorAnalysis} assigns every evaluation trajectory to a single outcome, which separates navigation failures (the agent never reaches the page containing the answer) from failures that occur after the relevant page has been found. The zero-shot model almost never completes a task. $\sim97\%$ of its trajectories end by it getting stuck in a loop. Behavior Cloning (BC) substantially reduces looping but leaves it as the dominant failure mode ($\sim62.5\%$). Behavior Cloning (TW) and m-GRPO both reach roughly 30\% success, yet each still loses $15.5\%-19.4\%$ trajectories to loops. Off-policy takeover produces the fewest unsupported answers among the methods (27). However, it also has the highest looping failure rate (36.6\%) among the non-BC methods.

\textsc{WebKD} passes the most tasks while getting stuck in a loop less than any other method ($\sim8.7\%$). Most of the gain comes from navigation. Counting trajectories that either pass or end in \emph{right page, wrong answer}, \textsc{WebKD} reaches the correct page 202 times, while no baseline does so more than 143 times. Because of this, its remaining errors move downstream, and \emph{right page, wrong answer} becomes its largest failure category ($\sim24.3\%$). This suggests that further improvements are more likely to come from better answer extraction once the relevant page is found than from better navigation. Task and judge defects remain roughly constant across methods (8--12 trajectories where labeled), indicating a small noise floor in the evaluation that does not depend on the agent.

\subsection{Where do the gains come from?}
\label{subsec:whereDogainsComeFrom}

\par\medskip
\noindent
\begin{minipage}[ht]{0.48\linewidth}
  \vspace{0pt}
  \centering
  \begin{tikzpicture}
  \begin{axis}[
    width=\linewidth, height=4.2cm,
    font=\scriptsize,
    ybar, bar width=0.55,
    ymin=0, ymax=30,
    ylabel={Gain (pp)},
    ytick={0,10,20},
    xtick={1,...,5},
    xticklabels={Lookup,Multi-hop,Count/list/sum,Verify/compare,Place order},
    x tick label style={font=\tiny,rotate=40,anchor=north east},
    tick label style={inner sep=1pt},
    axis lines*=left,
    tick align=outside,
    ymajorgrids, grid style={gray!15},
    enlarge x limits={abs=0.75},
    nodes near coords={+\pgfmathprintnumber[fixed,precision=1]{\pgfplotspointmeta}},
    nodes near coords style={font=\tiny, inner sep=1pt},
  ]
  \addplot[fill=after, draw=black!65, thin] coordinates {
    (1,8.7) (2,5.7) (3,9.5) (4,4.8) (5,25.9)
  };
  \end{axis}
  \end{tikzpicture}
  
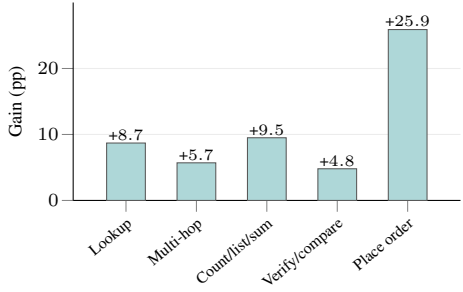
\captionof{figure}{Absolute gain by task type, in percentage points.}
  \label{fig:gain-by-task}
\end{minipage}%
\hfill
\begin{minipage}[ht]{0.48\linewidth}
  \vspace{0pt}
  Figure~\ref{fig:gain-by-task} breaks the overall improvement down by task
  type. Gains are positive in all five categories, but their size varies
  considerably. Placing an order improves the most, rising from 0\% to 25.9\%
  (+25.9\,pp), and it is the only category where performance started at zero.
  Counting/listing (+9.5\,pp, from 11.9\% to 21.4\%) and lookup (+8.7\,pp)
  show moderate gains, while multi-hop (+5.7\,pp) and verification
  (+4.8\,pp) improve the least. Because each category has only 9--29
  examples, these differences should be read with caution: only the gains on
  placing an order and counting/listing have 95\% confidence intervals that
  exclude zero.
\end{minipage}
\par\medskip

\subsection{{A Complete Worked Example}}
\label{app:qualitativeExample}

{This subsection gives the unabridged text of the step shown in Fig.~\ref{fig:rollout-edit}.}

\paragraph{Student, step 5.}
\begin{quote}\small\ttfamily
<think> We successfully purchased the Welch's item and are on the thank-you page showing code
B644DDC66C. Next, we need to search for the Mojito mix. I will click the search box (bid "15") to
bring up the search input. </think>\\
<plan> 1. Search for Mojito mix \quad - Click search box \quad - Type query \quad - Press Enter \quad
2. Open first result and buy \quad - Click first result link \quad - Click Buy Now and note
confirmation code \quad 3. Search for Italian Basil Pesto \quad [\ldots] \quad 5. Send all
confirmation codes to the user </plan>\\
<step> 2 </step>\\
<memory> I am clicking the search box to prepare for ordering the Mojito mix. </memory>\\
<action> click('15') </action>
\end{quote}

\paragraph{Raw teacher edit.}
\begin{quote}\small\ttfamily
<think> The page shows the order confirmation for Welch's Orange Pineapple Juice with code
B644DDC66C. I need now search for the second product, so I will navigate back to the shop home
page within the allowed SHOP environment. </think>\\
<plan> 1. Record Welch's confirmation code. 2. Navigate to shop home/search page. 3. Search for
Stirrings Simple Classic Cocktail Mix Mojito. 4. Open matching product, buy it, and record
confirmation code. 5. Navigate to shop home/search page. 6. Search for Italian Basil Pesto. 7. Open
matching product, buy it, and record confirmation code. 8. Send all confirmation codes to the
user. </plan>\\
<step> 2 </step>\\
<memory> Welch's Orange Pineapple Juice confirmation code: B644DDC66C. </memory>\\
<action> goto('http://localhost:5002/abc') </action>
\end{quote}

\paragraph{Style-conditioned teacher edit (the \name\ target), with span weights $v_s$ in brackets.}
\begin{quote}\small\ttfamily
<think> [$0.10$] The purchase confirmation page for Welch's Orange Pineapple Juice is displayed,
and the code shown is B644DDC66C. [$0.90$] I need to continue shopping for the next product, so
I will return to the shop search page. </think>\\
<plan> [$0.80$] 1. Record Welch's confirmation code \quad [$0.40$] 2. Search for Stirrings Simple
Classic Cocktail Mix Mojito \quad - Go to shop page \quad - Type query and press Enter \quad - Open
matching result \quad [$1.00$] - Click Buy Now and record code \quad [$0.70$] 3. Search for Italian
Basil Pesto \quad - Go to shop page \quad - Type query and press Enter \quad - Open matching result
\quad [$0.60$] - Click Buy Now and record code \quad [$0.50$] 4. Send all confirmation codes to the
user </plan>\\
<step> 2 </step>\\
<memory> [$0.30$] Welch's Orange Pineapple Juice confirmation code is B644DDC66C. [$0.20$] Next I
need to search for Stirrings Simple Classic Cocktail Mix Mojito. </memory>\\
<action> goto('http://localhost:5002/abc') </action>
\end{quote}
}

\section{Theoretical Intuition and Analysis}
\label{subsec:theoreticalIntuition}

\subsection{{What the Anchored Update Does}}

Write $p_i = \pi_\theta(y^*_i \mid \cdot)$ and $q_i = \pi_{\text{ref}}(y^*_i \mid \cdot)$ for the two
policies' probabilities of corpus token $y^*_i$, $\mathbf{p}^{(i)}$ for the student's full
next-token distribution at that position, $\mathbf{e}_{y^*_i}$ for the one-hot target, and collect
the two normalized coefficients acting on token $i$: $a_i = w_i / \sum_j w_j$ (imitation) and
$b_i = \beta \gamma_i / \sum_j \gamma_j$ (anchor). Because the pointwise anchor depends on the
position-$i$ logits $\mathbf{z}^{(i)}$ only through $\log p_i$, the gradient of Eq.~\ref{eq:kl}
at that position factorizes exactly:
\begin{equation}
\nabla_{\mathbf{z}^{(i)}} \mathcal{L}
\;=\;
\underbrace{\Big(a_i \;-\; b_i \big(1 - q_i/p_i\big)\Big)}_{\text{throttle } c_i}
\cdot
\underbrace{\big(\mathbf{p}^{(i)} - \mathbf{e}_{y^*_i}\big)}_{\text{SFT direction}}.
\label{eq:kl-grad}
\end{equation}
Eq.~\ref{eq:kl-grad} says the anchor never \emph{rotates} the update. Every anchored token
still moves along the SFT direction, {\color{black}toward the corpus token while $c_i > 0$ and away from it when $c_i < 0$}, and the anchor only rescales it through the
data-dependent {scalar coefficient $c_i$, which we call the throttle}:
\begin{itemize}
\item \textbf{At initialization the anchor is inert.} {Training starts from $\pi_{\text{ref}}$,
so $p_i = q_i$ and $c_i = a_i$, and the first update is exactly weighted SFT.} The penalty acts on
drift accumulated during training, never on the starting point.
\item \textbf{Below the reference, the anchor helps.} If interference drives $p_i$ under $q_i$
{(the student losing probability on text its prior found natural), then $q_i/p_i > 1$ and
$c_i > a_i$, so the anchor \emph{adds} force toward the corpus token, protecting prior competence
rather than fighting the corpus.}
\item \textbf{Above the reference, it throttles and can stop.} As $p_i$ climbs past $q_i$
the throttle falls monotonically toward $a_i - b_i$. When $b_i > a_i$ it {\color{black}vanishes at $p_i = q_i b_i/(b_i - a_i)$ and is negative beyond it, where the update pushes $p_i$ back down; that point lies below $1$, so the throttle actually stops the token, only when $b_i (1 - q_i) > a_i$ (Corollary~\ref{cor:beta})}, which is the content of Proposition~\ref{prop:target}.
\item \textbf{On $\langle\mathrm{A}\rangle$ and $\langle\mathrm{S}\rangle$, it is absent.} $\langle\mathrm{A}\rangle$ and $\langle\mathrm{S}\rangle$ carry
{$\gamma_i \approx 0$, so $b_i \approx 0$ and $c_i {\color{blue}\approx} a_i$ throughout, and the teacher's decision is
learned at full strength, all the way to the one-hot target.}
\end{itemize}

\begin{proposition}[Per-token target of the anchored objective]
\label{prop:target}
Fix a supervised position $i$, treat its logits as free parameters, and hold the context fixed.
The contribution of position $i$ to Eq.~\ref{eq:kl} is minimized at
\begin{equation}
p_i^\star \;=\;
\begin{cases}
\min\!\left\{1,\;\; q_i \cdot \dfrac{b_i}{b_i - a_i}\right\}, & b_i > a_i,\\[10pt]
1, & b_i \le a_i.
\end{cases}
\label{eq:pstar}
\end{equation}
(Proof in Appendix~\ref{app:theory}.)
\end{proposition}

That is, the anchor does not shrink the \emph{step} toward the corpus. It moves the
\emph{target}, from the one-hot corpus token to a bounded multiple of the {reference policy's} own
probability.

\begin{corollary}[Critical anchor strength]
\label{cor:beta}
The cap in Eq.~\ref{eq:pstar} exists iff
$\beta > \bar\beta_i \equiv (w_i / \gamma_i) \cdot \big(\textstyle\sum_j \gamma_j \big/ \sum_j w_j\big)$.
{Because $\bar\beta_i \propto w_i$, the cap {\color{black}appears} \emph{first} on {\color{black}the} spans the
action-informative weighting marked as least informative about the action.}
\end{corollary}
{The condition $\beta > \bar\beta_i$ makes the interior optimum of Eq.~\ref{eq:pstar}
finite. That optimum is below $1$, and so limits the token, only under the stronger
condition $q_i\, b_i / (b_i - a_i) < 1$, that is, $b_i (1 - q_i) > a_i$.} {\color{black}In anchor strength this is $\beta > \bar\beta_i / (1 - q_i)$, so the order in which caps bind also depends on $q_i$: among tokens with equal $q_i$ the least informative spans are capped first, but a span the reference already finds likely can stay uncapped at a $\beta$ that caps a more informative, less likely one.}


\subsection{{A Single-Round Relabeling Bound}}
\label{sec:bound}

The pipeline admits an imitation-learning guarantee in the style of DAgger \citep{ross2011dagger}, with the classical action oracle replaced by the two calls a black-box teacher actually answers: localize (P3) and correct (P4). Write $\rho(\pi)$ for the success rate of a policy on the task distribution and $\rho_{\mathrm{BC}}$ {\color{black}for the fraction of tasks whose logged seed rollout (P2) succeeded, an estimate of $\rho(\pi_{\theta_\text{BC}})$.} 
Three error rates describe one correction round{, each a population quantity over the seed's failed episodes}. $\varepsilon_{\mathrm{loc}} = \Pr[t' > t^\star]$ is the \emph{late}-localization rate of P3 (early misses are benign for the guarantee; they only spend edits). $\varepsilon_{\mathrm{corr}}$ is the teacher's own failure rate when it takes over as the agent from a still-recoverable localized state $h_{t'}$ and plays the episode out. 
{$\bar\varepsilon$ is the trained student's per-step disagreement, over the full structured response, with the relabeled process that replays the seed through step $t'-1$ and then follows the teacher, measured on that process's own history distribution{\color{black}, that is, $\bar\varepsilon = \frac{1}{T}\sum_{t=1}^{T}\delta_t$} (formal definitions in Appendix~\ref{app:bound}).}

\begin{theorem}[Single-round relabeling bound, informal]
\label{thm:bound}
Under the assumptions of Proposition~\ref{prop:tstar}, the student $\pi_{\theta_{\mathrm{WD}}}$ trained on the P1--P4 corpus satisfies
\begin{equation}
1 - \rho(\pi_{\theta_{\mathrm{WD}}})
\;\le\;
\underbrace{(1-\rho_{\mathrm{BC}})\big(\varepsilon_{\mathrm{loc}} + \varepsilon_{\mathrm{corr}}\big)}_{\text{\color{black}teacher term}}
\;+\;
T\,\bar\varepsilon,
\qquad
\bar\varepsilon \;\le\; \rho_{\mathrm{BC}}\,\varepsilon_{\mathrm{ret}} + (1-\rho_{\mathrm{BC}})\big(\varepsilon_{\mathrm{fit}} + \varepsilon_{\mathrm{rec}}\big),
\label{eq:bound}
\end{equation}
where $\varepsilon_{\mathrm{ret}}$ is {\color{black}disagreement with the seed's logged responses} on episodes the seed already solves (retention), $\varepsilon_{\mathrm{fit}}$ is disagreement {\color{black}with the relabeled process at the states the corpus supervises (the logged response before $t'$, and at $t'$ the teacher's response distribution, of which the corpus record is one draw)}, and $\varepsilon_{\mathrm{rec}}$ is disagreement at the post-correction states no corpus record covers. (Formal statement and proof in Appendix~\ref{app:bound}.)
\end{theorem}

Eq.~\ref{eq:bound} separates what the teacher limits from what the student limits. With a
perfect student ($\bar\varepsilon = 0$), the remaining failure is the fraction of seed failures
that the teacher either localizes too late or cannot repair itself. We estimate both error rates on the $15$ seed failures the
corpus is built from. Taking over at $t'$, the teacher solves $14$ of them. The one it fails was
localized at the first step, a state that is recoverable by assumption, so the failure counts
against correction rather than localization, giving $\varepsilon_{\mathrm{loc}} = 0/15$ and
$\varepsilon_{\mathrm{corr}} = 1/15$ on this pool. The teacher term is then
$0.70 \times 0.07 \approx 0.05$, well below the observed student failure rate of $0.61$, so the
teacher's errors are not the main limit. This is an estimate rather than a direct measurement of
the term, because the takeover failure rate comes from training-split states under one judge and
the seed failure rate from the test split under another. For reference, the teacher succeeds on
$0.71$ of test episodes in the same harness (Tab.~\ref{tab:main_results}).

The student term $T\bar\varepsilon$ is linear in $T$ because $\bar\varepsilon$ is evaluated on the
history distribution of the relabeled process, which the corpus supervises except at the
post-correction states counted in $\varepsilon_{\mathrm{rec}}$. This is where collecting the
corpus on the student's own states enters the bound. The bound concerns terminal failure, whereas
the $O(T^2\varepsilon)$ bound for cloning off-policy demonstrations
(\S\ref{subsec:BehaviorCloning}) concerns cumulative cost, so the two are not directly comparable.
{\color{black}Because the disagreement is measured on the full structured response, any rewording of a rationale counts as disagreement, so $T\bar\varepsilon$ is conservative and can exceed one. We use the bound for its structure, which separates the teacher's errors from the student's and places the student's error on the relabeled process's own histories, not as a numerical prediction of the student's failure rate.}

{
\subsection{{Premium Split, Rank Weighting, and the Anchor Budget}}
\label{sec:theory}

This subsection relates the per-token result of Proposition~\ref{prop:target} to the premium of
Definition~\ref{def:premium}. Under uniform SFT, the premium determines how the update splits
between rationale and action (Proposition~\ref{prop:split}). The rank map of
Eq.~\ref{eq:actw-map} orders the update optimally to first order {\color{black}in an additive twin with a common emission probability} (Theorem~\ref{thm:rank}). The
anchor limits how much premium each token can absorb, and the limit is largest on the spans that
the weighting marks as decisive (Corollary~\ref{cor:budget}). Proofs are in
Appendix~\ref{app:theory-update}.

\paragraph{Where the uniform update goes.}
Fix a set of corpus steps and score them under the seed $\pi_{\theta_\text{BC}}$, the policy
every run starts from. Write $\mathcal I_{\mathrm{rat}}$ for the positions of the rationale
tokens ($\langle\mathrm{T}^*\rangle$, $\langle\mathrm{P}^*\rangle$, $\langle\mathrm{M}^*\rangle$)
and $\mathcal I_{\mathrm{act}}$ for the positions of the action tokens across those steps, with
$n_{\mathrm{rat}} = |\mathcal I_{\mathrm{rat}}|$ and $n_{\mathrm{act}} = |\mathcal I_{\mathrm{act}}|$.
Write $\mathrm{PPL}_{\mathrm{rat}}$ and $\mathrm{PPL}_{\mathrm{act}}$ for the per-token
perplexity of Eq.~\ref{eq:ppl} pooled over each set, so that
$\log \mathrm{PPL}_C = \frac{1}{n_C}\sum_{i \in \mathcal I_C} \mathrm{CE}_i$, and
$G_C = \sum_{i \in \mathcal I_C} \|\nabla_{\mathbf z^{(i)}} \mathrm{CE}_i\|_2$ for the
logit-gradient mass that uniform SFT places on set $C$.

\begin{lemma}[{\color{black}Gradient norm is sandwiched by the surprisal}]
\label{lem:sandwich}
At every supervised position $i$,
$\;1 - e^{-\mathrm{CE}_i} \;\le\; \|\nabla_{\mathbf z^{(i)}} \mathrm{CE}_i\|_2 \;\le\; \sqrt2\,\big(1 - e^{-\mathrm{CE}_i}\big)$.
\end{lemma}

\begin{assumption}[Bounded surprisal]
\label{ass:cap}
Every per-token surprisal is bounded, $\mathrm{CE}_i \le \Lambda$ for some $\Lambda < \infty$.
\end{assumption}

\begin{proposition}[The premium fixes the split of the uniform update]
\label{prop:split}
(i) Restricted to $\mathcal I_{\mathrm{rat}} \cup \mathcal I_{\mathrm{act}}$, the share of
{\color{black}the token-pooled uniform loss (every supervised token weighted equally, rather than each step normalized by its length as in Eq.~\ref{eq:sft})} that the action tokens contribute at initialization is exactly
\begin{equation}
\frac{n_{\mathrm{act}} \log \mathrm{PPL}_{\mathrm{act}}}{n_{\mathrm{act}} \log \mathrm{PPL}_{\mathrm{act}} + n_{\mathrm{rat}} \log \mathrm{PPL}_{\mathrm{rat}}}.
\label{eq:loss-share}
\end{equation}
(ii) Under Assumption~\ref{ass:cap}, the action's share of the logit-gradient mass satisfies
\begin{equation}
\frac{G_{\mathrm{act}}}{G_{\mathrm{act}} + G_{\mathrm{rat}}} \;\le\; \frac{X}{X + Y},
\qquad
X = \sqrt2\, n_{\mathrm{act}} \log \mathrm{PPL}_{\mathrm{act}},
\qquad
Y = \frac{1 - e^{-\Lambda}}{\Lambda}\, n_{\mathrm{rat}} \log \mathrm{PPL}_{\mathrm{rat}}.
\label{eq:grad-share}
\end{equation}
\end{proposition}
}
{
\paragraph{Why the rank map is the right order.}
Proposition~\ref{prop:split} says how much of the update lands on the rationale. The weighting
of \S\ref{sec:actw} decides how that part is spent across segments. Fix a corpus
step with corrected action $\langle\mathrm{A}^*\rangle$ and rationale segments
$s_1, \ldots, s_n$ (Appendix~\ref{app:actw}), and write $\Delta_j$ for the score of
Eq.~\ref{eq:actw} on segment $s_j$.

\begin{assumption}[Additive twin]
\label{ass:additive}
For every subset $\mathcal U \subseteq \{1, \ldots, n\}$ of segments present in the rationale,
the student's log-probability of the corrected action is
$\log P_{\pi}(\langle\mathrm{A}^*\rangle \mid h, \mathcal U) = \alpha_0 + \sum_{j \in \mathcal U} \Delta_j$
for some constant $\alpha_0$, so the leave-one-out score of Eq.~\ref{eq:actw} recovers
$\Delta_j$ exactly.
\end{assumption}
\begin{assumption}[Independent emission]
\label{ass:emit}
At deployment the student writes segment $s_j$ into its own rationale independently with
probability $\kappa_j \in (0,1)$, and raising $\log \kappa_j$ by $\mathrm{d}u$ nats costs
$\mathrm{d}u$ nats of premium on that segment (Definition~\ref{def:premium}).
\end{assumption}

\begin{theorem}[Rank weighting is first-order optimal credit assignment]
\label{thm:rank}
Under Assumptions~\ref{ass:additive} and \ref{ass:emit}, the deployment probability of the
corrected action is
$P_\pi(\langle\mathrm{A}^*\rangle \mid h) = e^{\alpha_0} \prod_{j=1}^{n} \big(1 - \kappa_j + \kappa_j e^{\Delta_j}\big)$,
and the gain in $\log P_\pi(\langle\mathrm{A}^*\rangle \mid h)$ per nat of premium absorbed on
segment $s_j$ is
\begin{equation}
\mu_j \;=\; \frac{\partial \log P_\pi(\langle\mathrm{A}^*\rangle \mid h)}{\partial \log \kappa_j}
\;=\; \frac{\kappa_j\,(e^{\Delta_j} - 1)}{1 - \kappa_j + \kappa_j e^{\Delta_j}},
\label{eq:mu}
\end{equation}
which is strictly increasing in $\Delta_j$, zero at $\Delta_j = 0$, and negative for
$\Delta_j < 0$. If moreover $\kappa_j = \kappa$ for all segments of the step and a step on
$\mathcal L_{\mathrm{SW}}$ sheds premium on segment $s_j$ at a rate proportional to its weight
$v_j$ (Remark~\ref{rem:resample}), then for any fixed multiset of segment weights
$\{v_1, \ldots, v_n\}$ the first-order gain $\sum_j \mu_j v_j$ is maximized by assigning the
weights in the order of $\Delta_j$, and the maximizing assignment depends on the $\Delta_j$ only
through that order.
\end{theorem}
\begin{remark}[What the twin predicts]
\label{rem:rank}
{The twin makes two predictions.} First, segments with
$\Delta_j < 0$ have $\mu_j < 0$, so learning them lowers the deployment probability of the
corrected action. {The weighting of \S\ref{subsec:WebKDLossDetails}} {\color{black}does not mask them: it ranks them below the step's segments with $\Delta_j \ge 0$ and keeps their weights in $[\phi, 1]$; outside the dead zone a unique lowest-ranked segment reaches $\phi$}, and $42\%$ of segments fall in this class. The twin predicts that
masking them costs nothing and may help.
Second, the gain of $\mathcal L_\text{WD}$ over $\mathcal L_\text{CE}$ in
Table~\ref{tab:loss-ablation} should grow with the within-step spread $\sigma$ of $\Delta_j$,
since a step with $\sigma \approx 0$ has nothing to reorder, which is what the dead zone of
Eq.~\ref{eq:actw-map} already assumes.
\end{remark}
}

{
\paragraph{The anchor as a premium budget.}
Proposition~\ref{prop:target} gives the per-token target of the anchored objective. Read in the
units of Definition~\ref{def:premium}, it is a budget.

\begin{corollary}[Per-token premium budget]
\label{cor:budget}
Under the anchored objective of Eq.~\ref{eq:kl}, the premium the student can shed on a rationale
token $i$, its $\mathrm{CE}_i$ at initialization minus its $\mathrm{CE}_i$ at the per-token
optimum $p_i^\star$ of Eq.~\ref{eq:pstar}, {satisfies}
\begin{equation}
\log \frac{p_i^\star}{q_i} \;\le\; -\log\!\Big(1 - \frac{\bar\beta_i}{\beta}\Big)
\quad \text{whenever } \beta > \bar\beta_i,
\qquad
\bar\beta_i = \frac{w_i}{\gamma_i}\cdot\frac{\sum_j \gamma_j}{\sum_j w_j},
\label{eq:budget}
\end{equation}
with equality when the interior cap $q_i b_i/(b_i - a_i)$ is at most $1$ (when it exceeds $1$
the token is fit to $p_i^\star = 1$ and sheds its full $\mathrm{CE}_i$, which is below the
right-hand side), and is the full $\mathrm{CE}_i$ at initialization (the token may be fit to
$p_i^\star = 1$) when $\beta \le \bar\beta_i$. The budget decreases in $\beta$ and increases in $w_i$, so at a
fixed $\beta$ the spans the weighting marks as decisive receive the largest budget.
\end{corollary}

}

\section{{Proofs and Theoretical Remarks}}
\label{app:theory}

\subsection{The first fatal step: proof of Proposition~\ref{prop:tstar}}
\label{app:tstar}

Fix a task and a failed trajectory $\tau$ executed under the harness of \S\ref{subsec:background}. Write $s_{(t)}$ for the underlying browser state after the first $t$ actions of $\tau$, with $s_{(0)}$ the initial state, and recall the recoverability indicator $\nu_t$ of Definition~\ref{def:recover}. The assumptions of Proposition~\ref{prop:tstar}, stated in full:

\begin{assumption}[Finite budget]\label{as:budget}
{The harness enforces a step budget $T$, and an episode that issues a terminal action earlier enters an absorbing state with its reward fixed.}
\end{assumption}
\begin{assumption}[Determinism]\label{as:det}
$\mathbb T$, $\mathbb O$, and $f_{\mathrm{parse}}$ are deterministic, so the state reached by a given action sequence from $s_{(0)}$ is unique and replayable.
\end{assumption}
\begin{assumption}[Terminal binary reward]\label{as:reward}
$R \in \{0,1\}$ is assigned once, when the episode ends.
\end{assumption}
\begin{assumption}[Feasibility]\label{as:feasible}
$\nu_0 = 1$: the task is solvable from its initial state within the budget.
\end{assumption}

\begin{lemma}[Recoverability is absorbing]\label{lem:mono}
Under Assumptions~\ref{as:budget}--\ref{as:reward}, $\nu_t = 0$ implies $\nu_{t+1} = 0$ along $\tau$.
\end{lemma}
\begin{proof}
Suppose $\nu_{t+1} = 1$: some action sequence $(a'_1, \dots, a'_m)$ with $m \le T-(t+1)$ reaches $R=1$ from $s_{(t+1)}$. Since $s_{(t+1)} = \mathbb T(s_{(t)}, a_{t+1})$ deterministically, the sequence $(a_{t+1}, a'_1, \dots, a'_m)$ has length $m+1 \le T-t$ and reaches $R=1$ from $s_{(t)}$, so $\nu_t = 1$. The contrapositive gives the claim.
\end{proof}

\begin{proof}[Proof of Proposition~\ref{prop:tstar}]
{$\nu_0 = 1$ by Assumption~\ref{as:feasible}. At the episode's end $\nu = 0$. If the budget is exhausted, the only continuation within budget $0$ is the empty one, and it ends with the episode's own reward $R(\tau) = 0$. If a terminal action ended the episode early, the state is absorbing with $R = 0$ (Assumption~\ref{as:budget}). By Lemma~\ref{lem:mono} the sequence $\nu_0, \nu_1, \dots$ is non-increasing over $\{0,1\}$, so it crosses exactly once, and $t^\star = \min\{t \ge 1 : \nu_t = 0\}$ exists and is unique,} with $\nu_t = 1$ for all $t < t^\star$ and $\nu_t = 0$ for all $t \ge t^\star$. {For the final clause, $\nu_{t^\star-1} = 1$ supplies a within-budget success sequence from $s_{(t^\star-1)}$. Its first action $a$ leaves the task solvable in the remaining budget, so a recoverability-preserving action existed at step $t^\star$, and the executed $a_{t^\star}$ was not one, since it produced $\nu_{t^\star} = 0$.}
\end{proof}

\paragraph{Remarks.}
\emph{(i) {The role of the budget.}} Recoverability is defined \emph{within the remaining budget}, so even a diffuse failure with no single wrong click, an agent that dithers, loops, or re-reads pages, has a unique $t^\star$: the first step after which the shortest completion no longer fits in the budget. Without Assumption~\ref{as:budget} the proposition is false, since an agent on an unbounded horizon can wander indefinitely without failure ever becoming irreversible.
\emph{(ii) What $t^\star$ is and is not.} $t^\star$ is an interventional quantity, the earliest step after which \emph{no} policy could still succeed, which is what the localization instruction of Appendix~\ref{app:teacher} asks for. By Assumption~\ref{as:det} it is a property of the executed actions and the environment alone: attainability depends only on the browser state and the remaining budget, so rationale tokens influence which action a policy \emph{will} take but never whether success is still attainable. It need not coincide with the causal ``root cause'' a human would narrate (a misread page three steps earlier), 
which is why the {localization study of \S\ref{subsec:p3failureLocalizationDetails}} anchors annotators to the recoverability definition rather than to blame.
\emph{(iii) Live sites.} Assumption~\ref{as:det} holds exactly in a containerized benchmark and only approximately on the live web, where transitions are stochastic. There $\nu_t$ generalizes to the optimal success probability from $s_{(t)}$, monotonicity can fail (a lucky refresh can restore a lost option), {and $t^\star$ must be defined relative to a threshold on $\nu_t$. Nothing in the paper relies on this extension.}
\emph{(iv) Infeasible tasks.} {If $\nu_0 = 0$, the task is unsolvable from the start{\color{black}: Eq.~\ref{eq:tstar} gives $t^\star = 1$, but no recoverability-preserving action ever existed}, and no correction can lead to success.} Assumption~\ref{as:feasible} is thus a benchmark-design requirement, witnessed per task by any successful demonstration.

\subsection{Proof of Proposition~\ref{prop:target}}
\label{app:proof-target}

Restrict Eq.~\ref{eq:kl} to a single supervised position $i$ and write $p$ for
$p_i = \pi_\theta(y^*_i \mid \cdot)$, suppressing the index. Substituting
$e^{d_i} = q_i / p$ into $\mathrm{kl}_i = e^{d_i} - d_i - 1$, position $i$ contributes
\begin{equation}
L_i(p) \;=\; -\,a_i \log p \;+\; b_i \left(\frac{q_i}{p} + \log\frac{p}{q_i} - 1\right),
\end{equation}
with $a_i = w_i / \sum_j w_j$ and $b_i = \beta \gamma_i / \sum_j \gamma_j$ as in {Appendix~\ref{subsec:theoreticalIntuition}}.
Differentiating,
\begin{equation}
\frac{dL_i}{dp} \;=\; -\frac{a_i}{p} + b_i\!\left(-\frac{q_i}{p^2} + \frac{1}{p}\right)
\;=\; -\,\frac{c_i(p)}{p},
\qquad
c_i(p) \;=\; a_i - b_i\left(1 - \frac{q_i}{p}\right),
\end{equation}
recovering the throttle of Eq.~\ref{eq:kl-grad}. {For $b_i > 0$ the function $c_i(p)$ is
strictly decreasing in $p$, with $\lim_{p \to 0^+} c_i = +\infty$ and $c_i \to a_i - b_i$ as
$p \to \infty$, and for $b_i = 0$ it is the constant $a_i$.} If
$b_i \le a_i$, then $c_i(p) = (a_i - b_i) + b_i q_i / p > 0$ for every $p \in (0, 1]$
{(here $a_i > 0$, since $w_i \ge \phi > 0$)}, so $L_i$
is strictly decreasing on $(0,1]$ and the minimum sits at the boundary $p = 1$ {\color{black}(approached as the target logit grows, since finite logits give $p < 1$)}. If $b_i > a_i$,
solving $c_i(p) = 0$ gives the unique interior stationary point
\begin{equation}
p_i^\star \;=\; \frac{q_i\, b_i}{b_i - a_i},
\end{equation}
with $c_i > 0$ below it and $c_i < 0$ above it, so it is the minimum. Clipping at the boundary
yields Eq.~\ref{eq:pstar}. {\color{black}The minimization can also be checked without calculus, which is how the Lean proof does it. With $\hat p = q_i b_i/(b_i - a_i)$ and $b_i > a_i$, $L_i(p) - L_i(\hat p) = \frac{q_i b_i}{\hat p}\big[(\hat p/p - 1) - \log(\hat p/p)\big] \ge 0$ for every $p > 0$, because $\log x \le x - 1$. When $b_i \le a_i$, or when $\hat p > 1$, $L_i(p) - L_i(1) = (b_i - a_i)\log p + b_i q_i (1/p - 1) \ge 0$ on $(0, 1]$, by $\log p \le 0$ in the first case and by $\log p \ge 1 - 1/p$ with $b_i - a_i < q_i b_i$ in the second.} For the full-softmax statement, note
$\nabla_{\mathbf z}(-\log p) = \mathbf p - \mathbf e_y$ and
$\nabla_{\mathbf z} \log p = \mathbf e_y - \mathbf p$, while
$\partial \mathrm{kl}_i / \partial \log p = 1 - e^{d_i} = 1 - q_i/p$. Combining gives
Eq.~\ref{eq:kl-grad}, which vanishes iff $c_i = 0$ or $\mathbf p = \mathbf e_y$. {Under
gradient flow from the initialization $\mathbf p = \mathbf q$ (the reference's full distribution,
since $\theta_0 = \theta_{\text{ref}}$), every logit moves along the SFT direction scaled by $c_i$.
The target logit rises at rate $c_i(1 - p)$ and each off-target logit $z_v$ falls at rate
$c_i\, p_v$, in proportion to its current probability, so the gap between two off-target logits
$z_u - z_v$ moves at rate $c_i(p_v - p_u)$ and the more probable alternatives lose ground faster.
The off-target distribution therefore flattens rather than keeping the reference's shape, while
$p$ rises toward $p_i^\star$, where $c_i$ vanishes when $p_i^\star$ is interior.} \hfill $\square$

We note that the closed form exists \emph{because} the anchor is pointwise in $p$. A
full-distribution KL to $\pi_{\text{ref}}$ would couple every vocabulary coordinate and add a
second, non-collinear term to Eq.~\ref{eq:kl-grad}. The tabular {idealization} treats each position's logits as free and ignores
parameter sharing across positions. It characterizes the force each token contributes, not the
coupled optimum of the network. {\color{black}The clamp of $d_i$ to $[-20, 8]$ (Appendix~\ref{app:kl-impl}) is also outside the analysis. It is inactive for $q_i e^{-8} \le p_i \le q_i e^{20}$, so Proposition~\ref{prop:target} is unchanged for every token with $q_i \ge e^{-20}$ (below $q_i e^{-8}$ the clamp only removes the anchor's force, and $L_i$ stays decreasing there). For rarer tokens $L_i$ is again decreasing above $q_i e^{20}$, so its minimum over $(0,1]$ can move to $p_i = 1$ even when $p_i^\star < 1$.}



\subsection{A single-round imitation bound: proof of Theorem~\ref{thm:bound}}
\label{app:bound}

P1 is behavior cloning on teacher rollouts, whose error can compound quadratically in the horizon
\citep{ross2010efficient}. P2--P4 implement the oracle that DAgger uses to make it
linear \citep{ross2011dagger},
expert-quality labels at states drawn from $d_{\pi_{\theta_\text{BC}}}$, but in a setting the classical theory
does not cover: the expert is black-box, answers only in \emph{text}, runs for a single round
rather than DAgger's iterated no-regret scheme, and its labels can fail in ways an action
oracle's cannot (a mislocalized failure, a fluent reference to state that was never observed).
{This subsection accounts for these differences.} Throughout, Assumptions~\ref{as:budget}--\ref{as:feasible}
hold, $R$ is treated as ground truth (the LLM judge that implements it adds measurement noise
orthogonal to this analysis), and success rates $\rho(\cdot)$ average over the training-task
distribution and any policy stochasticity.

\paragraph{The relabeled process.} The comparator is the process $\tilde\pi$ that the corpus of
P2--P4 supervises. On a task whose {\color{black}logged} seed rollout succeeded, $\tilde\pi$ {\color{black}replays} that rollout
and nothing is edited. On a task whose seed rollout failed with
localization $t'$, $\tilde\pi$ replays the logged responses through step $t'-1$ and then hands
control to the teacher-as-agent of Appendix~\ref{app:teacher}, style-conditioned exactly as in P4,
which plays the episode out closed-loop from $h_{t'}$. {\color{black}All quantities are conditional on the logged P2 rollouts, one per task, so the replay is exact and $\tilde\pi$ is a policy of the task. Accordingly $\rho_{\mathrm{BC}}$ is the fraction of tasks whose logged rollout succeeded, which estimates $\rho(\pi_{\theta_\text{BC}})$ because P2 rolls out at temperature $0.01$ (Appendix~\ref{app:training}).}

\paragraph{Error terms.} All are population quantities, conditioned on failed seed rollouts where
indicated:
\begin{align*}
\varepsilon_{\mathrm{loc}} &= \Pr\big[\,t' > t^\star \,\big|\, \text{seed rollout failed}\,\big], && \text{late localization (P3)},\\
\varepsilon_{\mathrm{corr}} &= \Pr\big[\,\text{teacher-as-agent fails from } h_{t'} \,\big|\, \text{failed}, \; t' \le t^\star\,\big], && \text{{corrector capacity} (P4)},\\
\delta_t &= \mathbb{E}_{h \sim d^t_{\tilde\pi}}\Big[\mathrm{TV}\big(\pi_{\theta_{\mathrm{WD}}}(\cdot \mid h),\, \tilde\pi(\cdot \mid h)\big)\Big], && \text{{student disagreement}},
\end{align*}
where $d^t_{\tilde\pi}$ is the step-$t$ history distribution of the relabeled process and the
{total variation is over the full structured response $y$. Remark~iii explains why the action
alone does not suffice.} {\color{black}We write $\bar\varepsilon = \frac{1}{T}\sum_{t=1}^{T}\delta_t$ for the per-step disagreement of Theorem~\ref{thm:bound}.}

\begin{lemma}[Imitation on a deterministic harness]\label{lem:imit}
For any two policies $\pi, \tilde\pi$ on the harness of \S\ref{subsec:background},
$\rho(\tilde\pi) - \rho(\pi) \;\le\; \sum_{t=1}^{T} \mathbb{E}_{h \sim d^t_{\tilde\pi}}\big[\mathrm{TV}\big(\pi(\cdot\mid h),\, \tilde\pi(\cdot\mid h)\big)\big]$.
\end{lemma}
\begin{proof}
{\color{black}By Assumption~\ref{as:det} a history is determined by the task and the responses emitted so far. Write $hy$ for the history after response $y$ at $h$, and $V_\pi(h)$ for the probability that $\pi$ succeeds from $h$ within the remaining budget, so that $V_\pi(h) = \sum_y \pi(y \mid h)\, V_\pi(hy)$, with $V_\pi = R$ once the budget is spent. At every history,
\begin{equation*}
V_{\tilde\pi}(h) - V_\pi(h) \;=\; \sum_y \tilde\pi(y \mid h)\big(V_{\tilde\pi}(hy) - V_\pi(hy)\big) \;+\; \sum_y \big(\tilde\pi(y \mid h) - \pi(y \mid h)\big)\, V_\pi(hy).
\end{equation*}
Both rows sum to one and $0 \le V_\pi \le 1$, so the second sum equals $\sum_y \big(\tilde\pi(y \mid h) - \pi(y \mid h)\big)\big(V_\pi(hy) - \tfrac12\big) \le \tfrac12 \sum_y |\tilde\pi(y \mid h) - \pi(y \mid h)| = \mathrm{TV}(\pi(\cdot\mid h), \tilde\pi(\cdot\mid h))$. The first sum is the same gap one step later, averaged under $\tilde\pi$. Unrolling it from the initial history over the $T$ steps and averaging over tasks gives the claim. This is the coupling argument in recursive form: the second sum bounds the probability that a maximal coupling of $\pi(\cdot\mid h)$ and $\tilde\pi(\cdot\mid h)$ separates the two responses at $h$, and the first follows the coupled process along $\tilde\pi$ while they agree.}
\end{proof}

\begin{theorem}[Single-round relabeling bound, formal]\label{thm:bound-formal}
Under Assumptions~\ref{as:budget}--\ref{as:feasible},
\begin{equation}
1 - \rho(\pi_{\theta_{\mathrm{WD}}})
\;\le\;
(1 - \rho_{\mathrm{BC}})\big(\varepsilon_{\mathrm{loc}} + \varepsilon_{\mathrm{corr}}\big)
\;+\; \sum_{t=1}^{T} \delta_t .
\label{eq:bound-formal}
\end{equation}
\end{theorem}
\begin{proof}
First bound the relabeled process. On tasks whose {\color{black}logged} seed rollout succeeds, probability
$\rho_{\mathrm{BC}}$, $\tilde\pi$ {\color{black}replays} that rollout and succeeds, by
Assumption~\ref{as:det}. On a failed rollout, distinguish the localization outcomes. If
$t' > t^\star$, then $t'-1 \ge t^\star$, so $\nu_{t'-1} = 0$ by Proposition~\ref{prop:tstar} and
no continuation from $h_{t'}$ within the remaining budget can succeed: the episode is lost
regardless of the teacher. If $t' \le t^\star$, then $\nu_{t'-1} = 1$, the state is recoverable,
and the teacher-as-agent fails with probability $\varepsilon_{\mathrm{corr}}$ by definition.
Hence
\begin{equation*}
1 - \rho(\tilde\pi)
\;\le\;
(1-\rho_{\mathrm{BC}})\big(\varepsilon_{\mathrm{loc}} + (1-\varepsilon_{\mathrm{loc}})\,\varepsilon_{\mathrm{corr}}\big)
\;\le\;
(1-\rho_{\mathrm{BC}})\big(\varepsilon_{\mathrm{loc}} + \varepsilon_{\mathrm{corr}}\big).
\end{equation*}
{\color{black}The first inequality does not use the lateness of the localization: a late episode is counted as a failure, which bounds its failure probability whatever the teacher does there, so it holds for any localization. Proposition~\ref{prop:tstar} makes the count exact, since a late episode is then lost with probability one, and Remark~(i) uses this.}
Apply Lemma~\ref{lem:imit} with $\pi = \pi_{\theta_{\mathrm{WD}}}$ and combine.
\end{proof}

\begin{corollary}[Regime decomposition]\label{cor:regimes}
{Partition the states of $d^t_{\tilde\pi}$ into three cells. The first holds states of solved
seed episodes, where {\color{black}$\tilde\pi$ replays the logged rollout} and no record supervises. The second
holds states of failed episodes up to and including $t'$, where the corpus supervises exactly the
histories $\tilde\pi$ visits (replay records before $t'$, the edited record at $t'$). The third
holds post-correction states $t > t'$ of failed episodes, which no record covers.} Writing $\varepsilon_{\mathrm{ret}}$, $\varepsilon_{\mathrm{fit}}$,
$\varepsilon_{\mathrm{rec}}$ for the per-step average of
$\mathrm{TV}(\pi_{\theta_{\mathrm{WD}}}, \tilde\pi)$ on each cell respectively,
\begin{equation}
\sum_{t=1}^{T} \delta_t \;\le\; T\Big(\rho_{\mathrm{BC}}\,\varepsilon_{\mathrm{ret}} \;+\; (1-\rho_{\mathrm{BC}})\big(\varepsilon_{\mathrm{fit}} + \varepsilon_{\mathrm{rec}}\big)\Big),
\end{equation}
since each episode contributes at most $T$ steps to its cells. 
{\color{black}With $\bar\varepsilon = \frac{1}{T}\sum_t \delta_t$ this is the second inequality of Eq.~\ref{eq:bound}. The bound adds the two failed-episode cells; weighting each instead by its share of those episodes' steps makes it an equality.}
\end{corollary}

\paragraph{Remarks.}
\emph{(i) The DAgger correspondence.} $\varepsilon_{\mathrm{fit}}$ plays the role of DAgger's
training loss on the aggregated on-policy dataset, and $T\bar\varepsilon$ its linear-in-horizon
error term \citep{ross2011dagger}.
{The teacher {\color{black}term} can be measured directly. A takeover run, in which the teacher plays the
episode out from $h_{t'}$, reports the bracket $\varepsilon_{\mathrm{loc}} + (1 - \varepsilon_{\mathrm{loc}})\,\varepsilon_{\mathrm{corr}}$
of the proof directly, since a late localization fails with probability one, so
$(1-\rho_{\mathrm{BC}})$ times the measured takeover failure rate is a tighter {\color{black}teacher term} than the
$\varepsilon_{\mathrm{loc}} + \varepsilon_{\mathrm{corr}}$ that Eq.~\ref{eq:bound-formal} prints
and needs no separate estimate of $\varepsilon_{\mathrm{loc}}$.}
\emph{(ii) Teacher capacity {\color{black}sets the bound's floor}, and the degradation is graceful.} A weaker teacher
enters the bound three times, through $\rho_{\mathrm{BC}}$ (worse P1 demonstrations, themselves
subject to the off-policy $O(T^2\varepsilon)$ regime above), through $\varepsilon_{\mathrm{loc}}$,
and through $\varepsilon_{\mathrm{corr}}$, each linearly. The bound is one-directional: it caps
failure and does not preclude the student empirically exceeding the teacher-in-harness, but any
such gain is outside the guarantee.
\emph{(iii) Why the disagreement is over the full response.} $h_{t+1}$ contains $y_t$, so two
policies that always agree on the executed action but phrase their rationale differently still
diverge in history, and the coupling of Lemma~\ref{lem:imit} breaks. {This is the truncated-history argument of Appendix~\ref{subsec:scratchpadMemory} as it
appears in the bound. Rationale tokens are part of the harness state, which is why the method
supervises and anchors them.} Action-level agreement suffices exactly when rationale
differences never change a later decision, a policy-dependent property the anchor pushes toward
but does not certify.
\emph{(iv) The single-round residual.} $\varepsilon_{\mathrm{rec}}$ lives on the states the
teacher's correction steers into, which no record covers by design, since edits never execute
(P4ii). Two mechanisms keep it small: recovered states resemble the teacher-demonstration states
P1 already cloned, and the anchor {\color{black}limits how far $\pi_{\theta_{\mathrm{WD}}}$ moves from $\pi_{\theta_\text{BC}}$ on corpus tokens, which limits drift where the corpus is silent only indirectly, through shared parameters}. Iterating P2--P4 from $\pi_{\theta_{\mathrm{WD}}}$ would convert
{$\varepsilon_{\mathrm{rec}}$ into supervised coverage, which is DAgger's aggregation. } Likewise
$\varepsilon_{\mathrm{ret}}$ is the retention term the anchor of \S\ref{sec:kl} targets: on
solved episodes the reference behavior is the seed's own, so any drift there is pure loss. {Because the anchor is a pointwise, role-masked penalty on corpus tokens
(\S\ref{sec:kl}), it lowers $\varepsilon_{\mathrm{ret}}$ and $\varepsilon_{\mathrm{rec}}$ by
limiting drift, but it does not bound either through a distributional inequality.}
\emph{(v) What the deep-suffix records do.} Edited records at $k > t'$ sit at seed-visited states
that are unrecoverable whenever $t' = t^\star$ (Proposition~\ref{prop:tstar}), so they cannot
{rescue those episodes and do not appear in Eq.~\ref{eq:bound-formal}. They teach the policy how
to behave at bad states} and influence the bound only through generalization, that is, through
lowering $\varepsilon_{\mathrm{fit}}$ and $\varepsilon_{\mathrm{rec}}$ at nearby states.
{The grounding filter of Algorithm~\ref{alg:webkd} removes a kind of label error that an
action oracle cannot make, fluent references to state that was never observed, which would
otherwise corrupt these records.}

{
\subsection{Proofs for \S\ref{sec:theory}}
\label{app:theory-update}

\begin{proof}[Proof of Lemma~\ref{lem:sandwich}]
Write $\mathbf p = \mathbf p^{(i)}$ for the student's next-token distribution at position $i$,
$y = y^*_i$ for the target, and $p_y = e^{-\mathrm{CE}_i}$ for its probability. Since
$\mathrm{CE}_i = -\log \mathrm{softmax}(\mathbf z^{(i)})_y$, the gradient is
$\nabla_{\mathbf z^{(i)}} \mathrm{CE}_i = \mathbf p - \mathbf e_y$, so
$\|\nabla_{\mathbf z^{(i)}} \mathrm{CE}_i\|_2^2 = (1 - p_y)^2 + \sum_{v \ne y} p_v^2$.
The sum is non-negative, which gives the lower bound, and
$\sum_{v \ne y} p_v^2 \le \big(\sum_{v \ne y} p_v\big)^2 = (1 - p_y)^2$ because every
$p_v \ge 0$, which gives the upper bound.
\end{proof}

\begin{proof}[Proof of Proposition~\ref{prop:split}]
(i) By Eq.~\ref{eq:ppl} pooled over a set $C$,
$\log \mathrm{PPL}_C = \frac{1}{n_C} \sum_{i \in \mathcal I_C} \mathrm{CE}_i$, so
$\sum_{i \in \mathcal I_C} \mathrm{CE}_i = n_C \log \mathrm{PPL}_C$. The uniform loss restricted
to $\mathcal I_{\mathrm{rat}} \cup \mathcal I_{\mathrm{act}}$ is
$\frac{1}{n_{\mathrm{rat}} + n_{\mathrm{act}}}$ times the sum of the two, and the action's share
is the ratio in Eq.~\ref{eq:loss-share}. (ii) By Lemma~\ref{lem:sandwich} and
$1 - e^{-x} \le x$ for $x \ge 0$,
$G_{\mathrm{act}} \le \sqrt2 \sum_{i \in \mathcal I_{\mathrm{act}}} (1 - e^{-\mathrm{CE}_i}) \le \sqrt2 \sum_{i \in \mathcal I_{\mathrm{act}}} \mathrm{CE}_i = X$.
The function $x \mapsto 1 - e^{-x}$ is concave, so on $[0, \Lambda]$ it lies above its chord,
$1 - e^{-x} \ge (1 - e^{-\Lambda})\, x / \Lambda$. With Assumption~\ref{ass:cap} and
Lemma~\ref{lem:sandwich},
$G_{\mathrm{rat}} \ge \sum_{i \in \mathcal I_{\mathrm{rat}}} (1 - e^{-\mathrm{CE}_i}) \ge \frac{1 - e^{-\Lambda}}{\Lambda} \sum_{i \in \mathcal I_{\mathrm{rat}}} \mathrm{CE}_i = Y$.
Finally $t \mapsto t/(t + s)$ is increasing in $t$ and decreasing in $s$, so
$G_{\mathrm{act}}/(G_{\mathrm{act}} + G_{\mathrm{rat}}) \le G_{\mathrm{act}}/(G_{\mathrm{act}} + Y) \le X/(X + Y)$.
\end{proof}

\paragraph{Scope.} The proposition is about gradients with respect to the logits, similar to Proposition~\ref{prop:target}.

\begin{proof}[Proof of Theorem~\ref{thm:rank}]
Under Assumption~\ref{ass:emit} the set $\mathcal U$ of segments the student writes has
$\Pr[\mathcal U] = \prod_{j \in \mathcal U} \kappa_j \prod_{j \notin \mathcal U} (1 - \kappa_j)$,
and under Assumption~\ref{ass:additive}
\begin{equation*}
P_\pi(\langle\mathrm{A}^*\rangle \mid h)
= \sum_{\mathcal U} \Pr[\mathcal U]\, e^{\alpha_0} \prod_{j \in \mathcal U} e^{\Delta_j}
= e^{\alpha_0} \prod_{j=1}^{n} \big(1 - \kappa_j + \kappa_j e^{\Delta_j}\big),
\end{equation*}
since the sum over subsets factorizes coordinate by coordinate. Taking logarithms and
differentiating the $j$-th factor with respect to $\log \kappa_j$ gives Eq.~\ref{eq:mu}. Writing
$a = e^{\Delta_j}$, $\mu_j = \kappa_j (a - 1)/(1 - \kappa_j + \kappa_j a)$ has derivative in $a$
equal to $\big[\kappa_j (1 - \kappa_j + \kappa_j a) - \kappa_j^2 (a - 1)\big]$ over the square of
the denominator, which simplifies to $\kappa_j / (1 - \kappa_j + \kappa_j a)^2 > 0$, so $\mu_j$
is strictly increasing in $\Delta_j$. It vanishes at $a = 1$ and is negative for $a < 1$.

For the ordering claim, the rate hypothesis says a step on $\mathcal L_{\mathrm{SW}}$ raises
$\log \kappa_j$ by $\eta\, v_j$ for a common $\eta > 0$. To first order the change in
$\log P_\pi(\langle\mathrm{A}^*\rangle \mid h)$ is then $\eta \sum_j \mu_j v_j$ by
Eq.~\ref{eq:mu}. With $\kappa_j = \kappa$ for all $j$, $\mu_j$ is an increasing function of
$\Delta_j$ alone, so the $\mu_j$ and the $\Delta_j$ have the same order. The rearrangement
inequality states that for two finite real sequences the sum of products is maximized when both
are sorted the same way, so among all assignments of a fixed multiset $\{v_j\}$ to the segments,
$\sum_j \mu_j v_j$ is maximized by assigning the weights in the order of $\Delta_j$. The
maximizing assignment is a function of the order of the $\Delta_j$ alone, which is what
Eq.~\ref{eq:actw-map} computes, up to ties, which the average rank resolves without changing the
maximum.
\end{proof}

\paragraph{Scope.} Assumption~\ref{ass:additive} makes the leave-one-out score exact and
removes interactions among segments, so a segment that matters only jointly with another is
scored as inert. Assumption~\ref{ass:emit} treats the segments as independently emitted and
prices a nat of log-probability on a segment as a nat of premium, which is
Definition~\ref{def:premium} applied to that segment. The rate hypothesis holds in logit space
when the rationale tokens of a step have equal gradient norm at initialization {\color{black} and its segments have equal token length}. A logit-space
step of size $\eta$ on $\mathcal L_{\mathrm{SW}}$ at a token $i$ with weight $w_i$ moves the
logits by $-\eta\, w_i (\mathbf p - \mathbf e_y)$ up to the common normalizer $\sum_k w_k$, and
since $\nabla_{\mathbf z^{(i)}} \log p_i = \mathbf e_y - \mathbf p$, it raises $\log p_i$ by
$\eta\, w_i \|\mathbf p - \mathbf e_y\|^2$ to first order. {\color{black}Summed over the tokens of $s_j$, it raises $\log \kappa_j$ by $\eta\, v_j\, g_j$ up to the normalizer, where $g_j = \sum_{i \in s_j} \|\mathbf p^{(i)} - \mathbf e_{y_i}\|^2$, so the rate is proportional to $v_j$ only when the $g_j$ agree.} Equal surprisal across the step's
rationale tokens makes these squared norms equal within a factor of two by
Lemma~\ref{lem:sandwich}, and equal when the off-target mass is spread alike. {\color{black}In general the first-order gain is $\eta \sum_j \mu_j\, g_j\, v_j$ and the optimal order is by $\mu_j g_j$. Here $\mu_j$ also depends on $\kappa_j$: by Eq.~\ref{eq:mu} its magnitude grows with $\kappa_j$, so it increases with $\kappa_j$ when $\Delta_j > 0$ and decreases when $\Delta_j < 0$.}

\begin{proof}[Proof of Corollary~\ref{cor:budget}]
By Proposition~\ref{prop:target}, if $b_i > a_i$ the per-token optimum is
$p_i^\star = \min\{1, q_i\, b_i / (b_i - a_i)\}$, so
$\log(p_i^\star / q_i) \le \log\big(b_i/(b_i - a_i)\big) = -\log(1 - a_i/b_i)$. With
$a_i = w_i / \sum_j w_j$ and $b_i = \beta \gamma_i / \sum_j \gamma_j$,
$a_i / b_i = \frac{w_i}{\gamma_i} \cdot \frac{\sum_j \gamma_j}{\sum_j w_j} \cdot \frac{1}{\beta} = \bar\beta_i / \beta$,
which is Eq.~\ref{eq:budget}, and $b_i > a_i$ is $\beta > \bar\beta_i$
(Corollary~\ref{cor:beta}). Training starts at $p_i = q_i$ (\S\ref{sec:kl}), so the surprisal
shed on token $i$ between initialization and the per-token optimum is
$\log p_i^\star - \log q_i$. If $b_i \le a_i$ the optimum is $p_i^\star = 1$ and the shed
surprisal is the full $\mathrm{CE}_i$ at initialization. The budget $-\log(1 - \bar\beta_i/\beta)$
is decreasing in $\beta$ and increasing in $\bar\beta_i \propto w_i$. On tokens with
$\gamma_i \approx 0$, $b_i \approx 0 < a_i$ and there is no cap.
\end{proof}
}

\section*{Declaration of LLM Usage}

The authors acknowledge the use of LLMs to polish the draft, {to} brainstorm ideas and research directions, to orchestrate experiments, {to} {\color{black}formalize and check the proofs in Lean}, and {to} review code. We take full responsibility for the paper's content and correctness.

\section*{Reproducibility Statement}

We have taken several steps to ensure our results are reproducible. The complete source code is included in the supplementary material and will be released publicly upon acceptance. It comes with a README describing environment setup, dependencies, and each component. {\color{black}Every proposition, lemma, corollary, and theorem of the theory appendix is formalized in Lean 4 on finite models of the setting, and the formalization is included in the supplementary material. Its README maps each result to its Lean declaration and states the modeling choices and what is trusted.} All datasets are publicly available and have been cited.

\end{document}